%% file: main.tex
\documentclass[logosize=10.5,showdate=false]{phai}
\usepackage{fontawesome5}
\usepackage{placeins}
\input{registry_counts}
\definecolor{aitonomyblue}{HTML}{7DA2FF}

\newcommand{\metadataicon}[1]{\makebox[1.25em][l]{\raisebox{-0.1ex}{\normalfont\fontsize{8.8}{8.8}\selectfont\faIcon{#1}}}}

\newcommand{\linksep}{\hspace{0.75em}\textcolor{phaifaint}{\textbar}\hspace{0.75em}}
\newcommand{\paperlinks}{%
  \vspace{2.4mm}
  {\sffamily\mdseries\fontsize{10}{12}\selectfont
   \href{https://aitonomy.org/projects/scienceide}{\faIcon{globe}\,Project}\linksep
   \href{https://github.com/aitofound/ScienceIDE}{\faIcon{github}\,GitHub}\linksep
   \href{https://huggingface.co/collections/AItonomy/scienceide-model-series}{\faIcon{robot}\,Model}\par}}
\makeatletter
\newcommand{\summarypanellabel}[2]{%
  \begingroup\protected@edef\@currentlabel{\thefigure#1}\label{#2}\endgroup}
\makeatother

\renewcommand*\phailockup[1][15]{%
   {\ttfamily\mdseries\fontsize{#1}{#1}\selectfont
   \raisebox{-0.16ex}{\includegraphics[height=9pt,trim=0 0 910 0,clip]{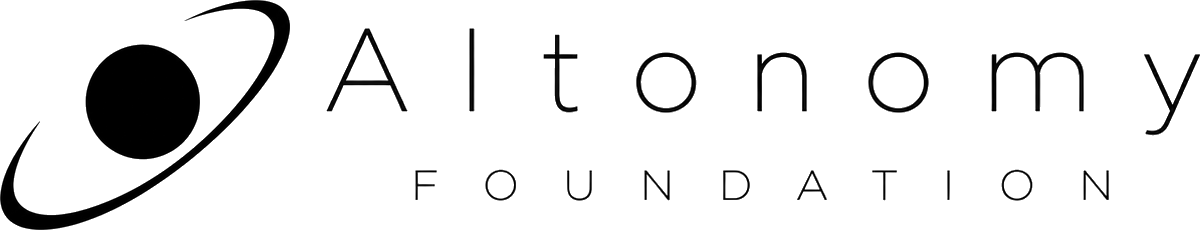}}\hspace{0.35em}%
   \textcolor{phaiink}{AI}\textcolor{aitonomyblue}{tonomy}%
   \textcolor{phaiink}{\space Foundation}}}

\title{ScienceIDE:\\Turning World’s Scientific Codebase into Agent Learnable Environments}

\authorheading{Authors}
\authors{%
  Hejia~Geng\aff{1,2*}\quad
  Zesen~Huang\aff{3*}\quad
  Haoyang~Li\aff{1}\quad
  Wenbin~Li\aff{1}\quad
  Koutian~Wu\aff{4}\quad
  Zihan~Zhou\aff{5}\quad
  Yuanbo~Pang\aff{6}\quad
  Weihao~Liu\aff{7}\quad
  Zigong~Xu\aff{8}\quad
  Zhiping~Li\aff{5}\quad
  Zongzheng~Zhang\aff{1}\quad
  Chuanfei~Dong\aff{9}\quad
  Jiankai~Sun\aff{10}\quad
  Tianzhe~Zheng\aff{8}\quad
  Fengyu~Xie\aff{1}\quad
  Yue~Ma\aff{11}\quad
  Yueheng~Shi\aff{10}\quad
  Junkai~Wang\aff{1}\quad
  Tong~Xie\aff{12}\quad
  Zonglin~Di\aff{6,13}\quad
  Xianrong~Liu\aff{12}\quad
  Qucheng~Gao\aff{14}\quad
  Yimin~Liu\aff{15}\quad
  Jiaming~Pan\aff{7}\quad
  Sheng~Huang\aff{9}\quad
  Xiao-Han~Ma\aff{16}\quad
  Lanqing~Yuan\aff{17}\quad
  Zhenlin~Zhu\aff{3}\quad
  Ziang~Liu\aff{18}\quad
  Ziyang~Xu\aff{19}\quad
  Kangkai~Liang\aff{11}\quad
  Jiayi~Xian\aff{20}\quad
  Zehong~Zhao\aff{11}\quad
  Taigao~Ma\aff{7}\quad
  Fuming~Chang\aff{7}\quad
  Yuanzhe~Hu\aff{25}\quad
  Liuwei~Xu\aff{8}\quad
  Xianzhe~Tang\aff{9}\quad
  Jingxu~Xie\aff{6}\quad
  Peijin~Zhang\aff{21}\quad
  Qiang Gao\aff{1}\quad
  Chengyi~Xing\aff{10}\quad
  Zhe~Zhao\aff{10}\quad
  Xi~Wang\aff{22}\quad
  Leo~Wang\aff{23}\quad
  Fang~Wu\aff{10}\quad
  Jiacheng~Ge\aff{22}\quad
  Weixuan~Liao\aff{9}\quad
  Jungang~Chen\aff{1}\quad
  Shuozhe~Li\aff{4}\quad
  Yi~Daniel~Lu\aff{26}\quad
  Xu~Pan\aff{26}\quad
  Shuhang~Lin\aff{27}\quad
  Peihan~Tu\aff{28}\quad
  Nan~Liu\aff{9}\quad
  Zelin~Tan\aff{1}\quad
  Yunze~Xiao\aff{29}\quad
  Hao~Xiao\aff{1}\quad
  Yaopeng~Xing\aff{24}\quad
  Xing~Meng\aff{24}\quad
  Zhenfei~Yin\aff{1,2\textdagger}\quad
  Yingcheng~Wu\aff{1,10\textdagger}\quad
  Ling~Yang\aff{1,5\textdagger}%
  \par
  \vspace{1mm}
  {\sffamily\mdseries\fontsize{8.6}{10.6}\selectfont\color{phaimuted}
   \textsuperscript{*}Equal contribution.\quad
   \textsuperscript{\textdagger}Corresponding author.\par}%
  \paperlinks
}

\sponsors{\mbox{PhAI-Labs\quad Qwen}}

\checkdata[\metadataicon{envelope}Email]{\href{mailto:team@aitonomy.org}{\texttt{team@aitonomy.org}}; \href{mailto:yang@phai-labs.com}{\texttt{yang@phai-labs.com}}}

\runningtitle{ScienceIDE}

\begin{document}
\maketitle

\begin{abstract}
Scientific code repositories encode decades of human knowledge in
executable models, methods, and tools. Yet fragmented toolchains,
implicit domain conventions, and specialized correctness criteria
make this knowledge difficult to convert into reliable learning
experience---a challenge we call the \emph{scientific experience
bottleneck}.
We introduce \textbf{ScienceIDE}, infrastructure for turning the
world's scientific code into programmable environments for scientific
agents. Guided by expert-defined scientific cases and acceptance
criteria, agents transform repositories into executable environments
that support task generation, execution, and scientific verification.
These environments provide a shared foundation for supervised
fine-tuning, reinforcement learning, and evaluation.
Using verified interaction trajectories, we train
\textbf{PhAI-IDE-72B/9B/4B}. The model family
shows gains in held-out scientific-code repair and across selected
general-purpose benchmarks in code, reasoning, and knowledge,
providing evidence of positive transfer from scientific experience
to broader capabilities.
ScienceIDE lays the foundation for an integrated workspace for
agent learning and scientific practice, making humanity's scientific
software a shared substrate for developing scientific intelligence.
\end{abstract}

\clearpage
\begin{figure}[!ht]
  \centering
  \includegraphics[width=\linewidth]{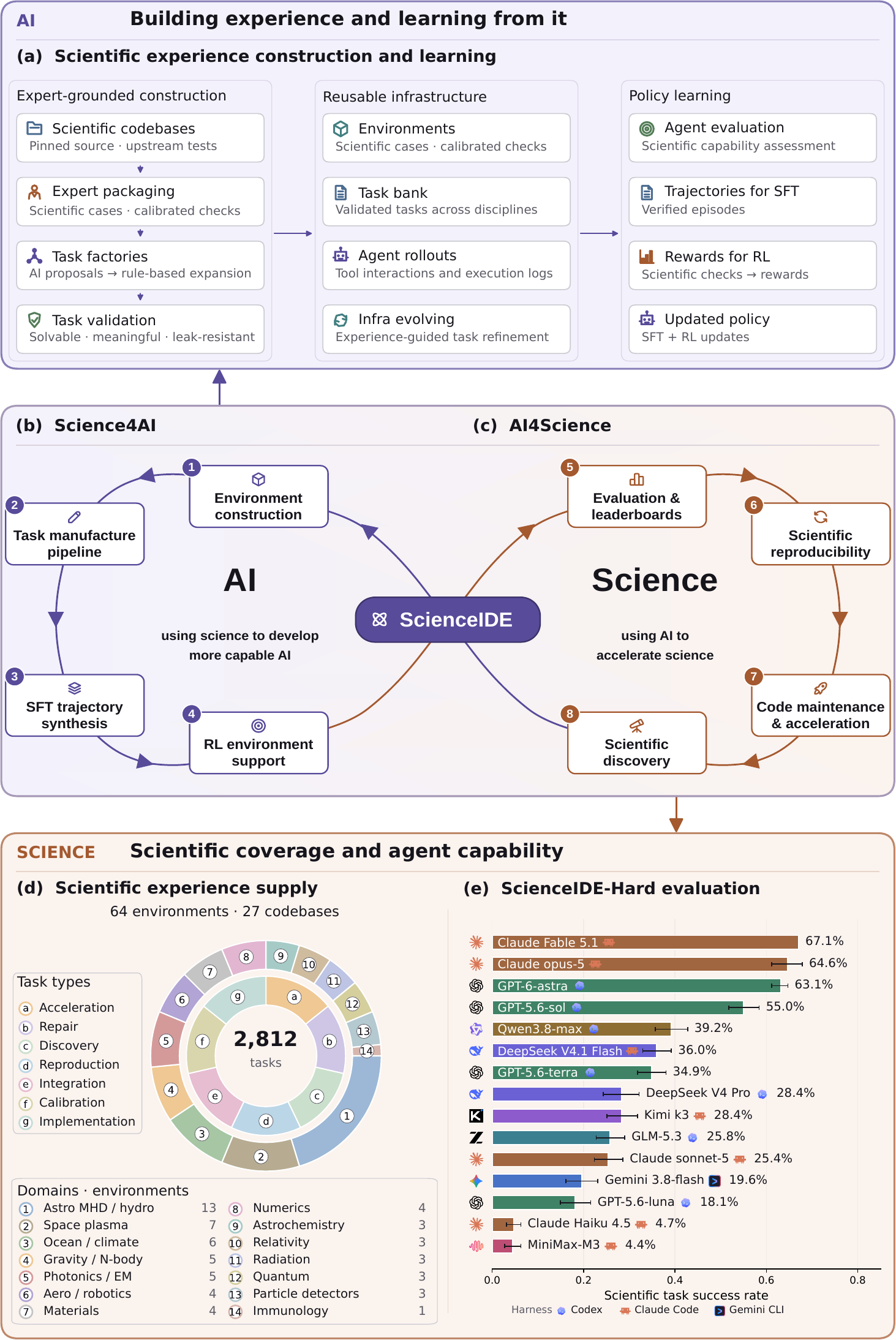}
  \caption{\textbf{ScienceIDE connects scientific work with agent learning.}
  (a) Expert-grounded construction provides reusable infrastructure for agent
  evaluation and learning.
  (b) Science4AI turns scientific experience into more capable agents.
  (c) AI4Science applies agents to scientific work and discovery.
  (d) Environment coverage by domain (outer ring) and task taxonomy
  (equal inner sectors, not task frequencies).
  (e) Model--harness performance on ScienceIDE-Hard; icons identify harnesses,
  and intervals summarize repeated runs.}
  \label{fig:summary}
  \summarypanellabel{b,c}{fig:teaser}
  \summarypanellabel{d}{fig:registry}
  \summarypanellabel{e}{fig:leaderboard}
  \summarypanellabel{a}{fig:summary-data}
  \summarypanellabel{a}{fig:summary-sft}
  \summarypanellabel{a}{fig:summary-rl}
\end{figure}
\clearpage

\input{intro}

\input{overview}       
\input{environments}
\input{sciencelobe}
\input{posttraining}
\input{experiments}    
\input{results}
\input{relatedwork}
\input{limitations}
\input{conclusion}

\bibliography{references,environments}

\supplementary
\input{organizations}
\input{appendix_compact_science}
\input{appendix_compact_evaluation}
\input{agent_diagnostics_appendix}
\input{appendix_compact_learning}
\input{environment_sources_compact}

\end{document}

%% file: registry_counts.tex
\newcommand{\RegistryEnvs}{64}
\newcommand{\RegistryTasks}{2{,}812}
\newcommand{\RegistrySources}{27}

%% file: intro.tex
\section{Introduction}

Language-model capability grows with the experience available for learning:
text supports chat intelligence through human knowledge and instruction
following \citep{ouyang2022instructgpt, christiano-rlhf}, while repositories,
compilers, and tests make coding actions observable and outcomes checkable, letting models act, observe consequences, and learn from failures \citep{swe-bench, pan2024swegym, jain2025r2egym,
yang2025swesmith}. Such experience also enables reinforcement learning with
verifiable rewards \citep{deepseek-r1, swe-rl}. However, scientific
intelligence remains challenging: whereas coding tasks usually specify a
problem, goal, and acceptance test, discovery-oriented work asks agents to identify worthwhile questions, test hypotheses through
interventions, and transfer lessons from evidence
\citep{silver2025experience, sutton2018rl}. Science concentrates these
demands in sustained interactions with code, data, and simulators, where
validated numerical models supply externally checked training signals.

Turning this opportunity into training experience exposes a scientific
experience bottleneck. Heterogeneous toolchains and configurations make scientific repositories difficult to execute reproducibly;
verification depends on physical quantities, numerical tolerances, and
unwritten conventions; and runnable code must become meaningful
tasks with behavioral validation, graded feedback, and recorded interactions
before it can support learning. Thus papers, repositories, and datasets do not automatically yield trainable environments. Scientific coding and
reproducibility benchmarks \citep{tian2024scicode, chen2024scienceagentbench,
siegel2024corebench, xu2026swebenchscience} evaluate agents but do not supply
the production infrastructure that expands experience across repositories
and task families.

The bottleneck is sharpest where the payoff is largest. Moving scientific
codes to accelerators still proceeds one hand-written port at a time
\citep{rossazza2026gpluto, an2025batsrusgpu}, each bound to a vendor
toolchain \citep{romeo2026portability} and to double precision as a default
rather than a derived requirement \citep{duben2014inexact,
duben2015programmable}. ScienceIDE fixes the contract that already exists in a code's own tests and expert tolerances as executable checks per environment and decouples it from task authoring, so any task, from injecting a defect to porting an expensive
path, is graded by whether it recovers the science within tolerance. The
registry so far holds 2,515 repair, 295 implementation, and two acceleration
tasks. On the hard subset, budget exhaustion reaches a task-balanced 37.3\%
(\S\ref{sec:experiments}); in the Fable/Astra trace review, reference-convention
mismatches account for most task-balanced failures
(Appendix~\ref{app:trace-contrasts}). These findings illustrate the long-horizon
scientific work this infrastructure exists to train.

ScienceIDE is open infrastructure that makes scientific expertise
reusable. Domain experts define executable repository environments, scientific
cases, and calibrated checking policies, while AI agents help instantiate them;
environment-specific factories reuse these decisions to generate valid tasks.
Execution and scientific verification accept candidate tasks, and agent
interactions over them yield graded trajectories; common interfaces provide
supervised fine-tuning (SFT) data,
online reinforcement-learning (RL) rewards, and evaluation with tasks or
environments held out from training. Expert effort therefore concentrates on
scientific boundaries and new objectives rather than each generated task. One environment thus
serves repair, implementation, reproduction, and acceleration alike. An open registry lets users train and evaluate agents on a chosen domain and return them to scientific work, where their interactions supply new tasks (Figure~\ref{fig:teaser}).

%% file: overview.tex
\section{ScienceIDE}
\label{sec:overview}

\begin{figure}[!htbp]
  \centering
  \includegraphics[width=\linewidth]{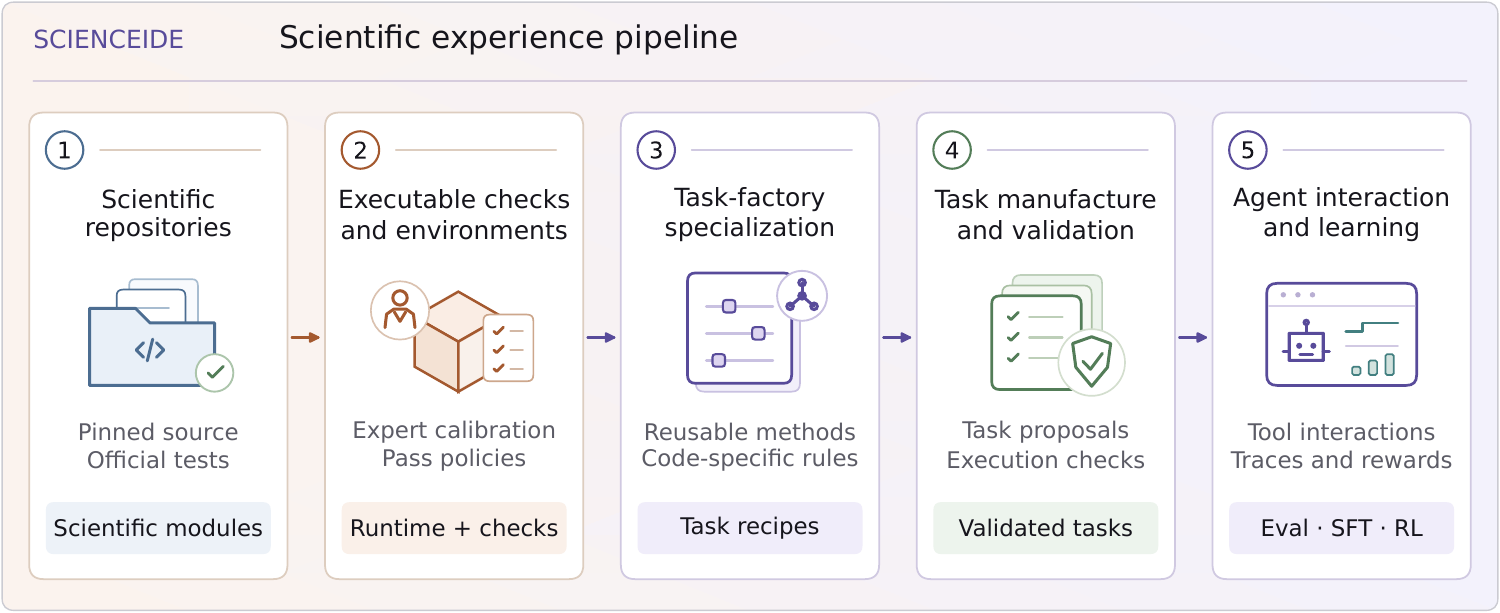}
  \caption{\textbf{ScienceIDE technical overview.}
  A versioned repository is packaged with a runtime and scientific checks.
  Environment-specific factories propose candidates; validation admits tasks
  whose episodes support evaluation, SFT, and RL.}
  \label{fig:arch}
\end{figure}

ScienceIDE turns scientific expertise into reusable agent experience.
Experts define scientific responsibilities and acceptance criteria;
executable environments preserve these decisions, task factories generate
challenges, and agent interactions provide evidence for evaluation and learning
(Figure~\ref{fig:arch}).

The construction unit is a scientific \emph{module} within a versioned
\emph{codebase}. A module owns a coherent scientific responsibility and
executable coverage, rather than simply grouping related files.
An \emph{environment} packages an approved module with its runtime and
scientific checks. Scaling adds modules and codebases while reusing their
acceptance criteria across task families.

A \emph{factory} specializes authoring procedures to an environment.
Its validated \emph{tasks} specify an initial workspace, a deliverable,
and a verifier. An \emph{episode} is one agent interaction with a task,
producing artifacts, a trajectory, and measured outcomes. The following
stages explain how these objects are constructed and connected.

%% file: environments.tex
\subsection{Compiling scientific expertise into executable environments}
\label{sec:environments}
\label{sec:envrecon}
\label{sec:reproducibility}

\begin{figure}[!htbp]
  \centering
  \includegraphics[width=\linewidth]{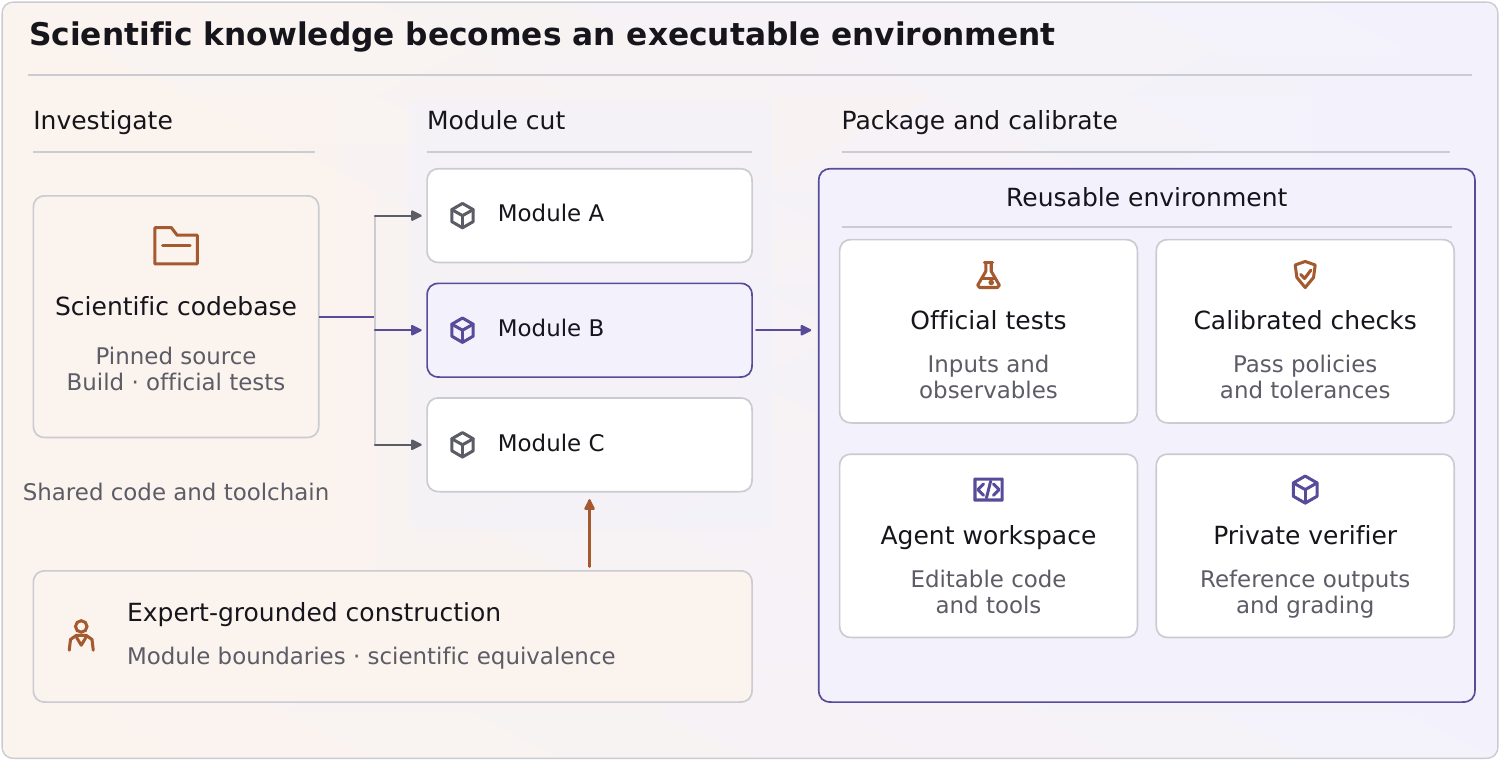}
  \caption{\textbf{From a codebase to an executable scientific contract.}
  Experts define module boundaries and scientific equivalence. Each selected
  module packages official tests with calibrated pass policies, an editable
  agent workspace, and a separate private verifier. These assets are reused
  across task families rather than reconstructed for each task.}
  \label{fig:expertai}
\end{figure}

Construction begins with production scientific software
(Figure~\ref{fig:expertai}). An agent inspects a pinned upstream revision,
dependencies, licence, and build assumptions, then builds the source and
runs official tests and examples. These runs expose output formats,
numerical variability, expensive paths, and execution hazards that inform
the environment boundary.

The agent proposes modules defined by scientific responsibility and
executable coverage. Each module identifies its inputs and outputs,
algorithm stages, owned implementation paths, excluded responsibilities,
and supporting tests. Shared solvers and toolchains may support several
modules without erasing their scientific distinctions.
A domain expert reviews the decomposition and coverage, allowing one
repository to contribute several environments with traceable source identity.

The agent implements the environment-specific experiment adapter and
proposes scientific outputs for review; the curator owns module boundaries
and the equivalence contract. The registry and CLI enforce structure and
track provenance without choosing scientific observables.
An approved module is packaged with an editable workspace, checks,
and a private verifier. Source pins, dependencies, and observed hazards
remain attached to this reusable runtime.

\input{check_design_method}

\subsection{Specializing reusable methods into task factories}
\label{sec:factory}

\begin{figure}[!htbp]
  \centering
  \includegraphics[width=\linewidth]{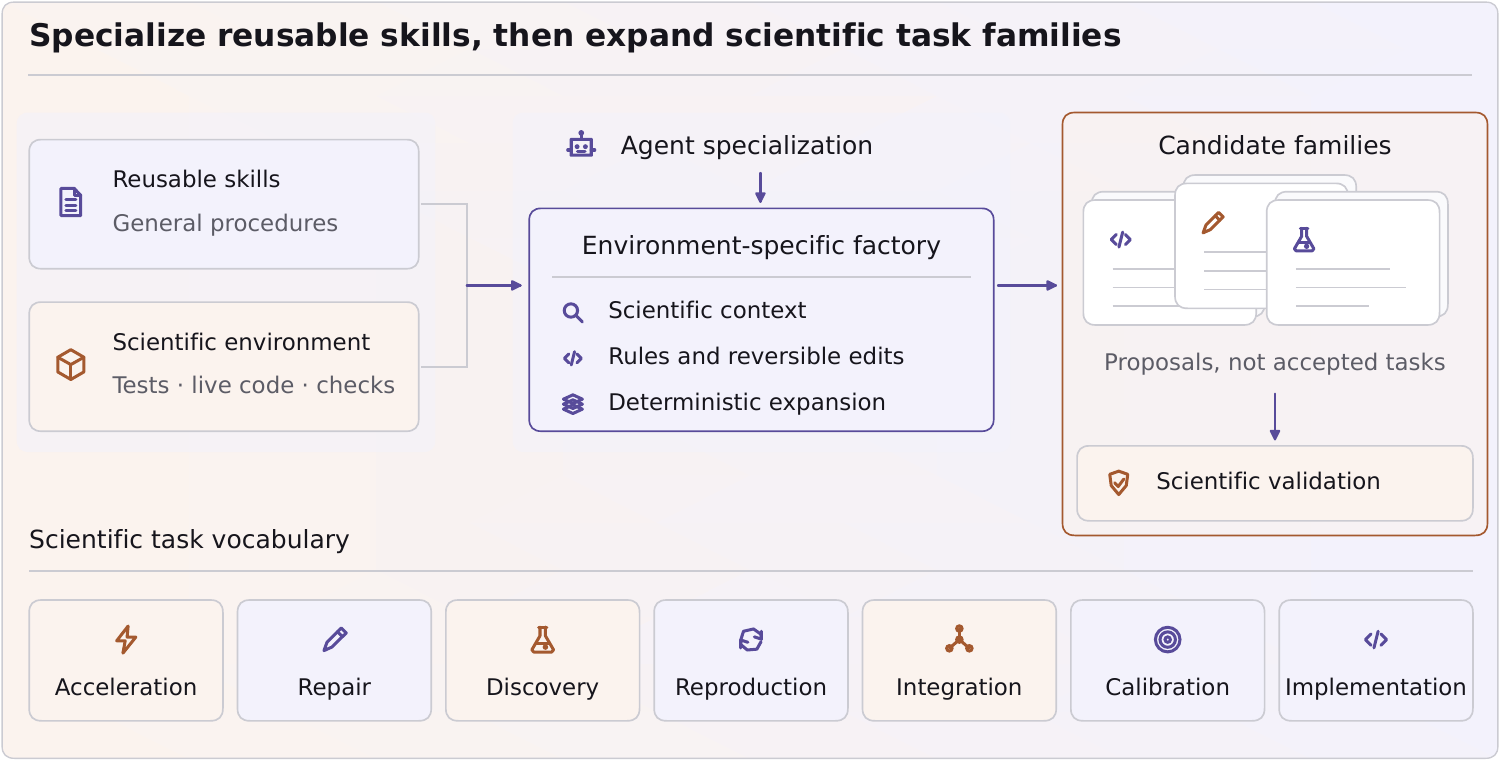}
  \caption{\textbf{General skills become environment-specific task factories.}
  An agent adapts reusable procedures to a scientific environment; local
  rules expand reproducible candidates, which become tasks only after later
  scientific validation. The seven categories describe the authoring
  taxonomy, not the measured coverage of deployed generators.}
  \label{fig:skillsloop}
\end{figure}

A factory combines reusable authoring procedures with an environment's
scientific context (Figure~\ref{fig:skillsloop}). Shared procedures handle
reversible edits, execution, and artifact assembly; local rules identify
active paths, cases, build recipes, and meaningful transformations.
The packaged module map, observables, tolerance evidence, and known
limitations constrain candidate generation.

The agent proposes semantic edit sites and objectives, while the curator
and domain expert retain decisions about observables, equivalence, and
acceptance. Deterministic procedures expand approved rules into mutation
or excision candidates; the validation stage determines which become tasks.
For example, a LAPS environment can reuse its scientific checks for both
defect repair and reconstruction of a missing timestep routine.
The requested work changes, while the numerical behavior to be recovered
remains defined by the same environment.

The authoring taxonomy has seven categories: Acceleration, Repair,
Discovery, Reproduction, Integration, Calibration, and Implementation.
Repair and Implementation expose automatable candidate expansion through
reversible mutation and excision. The other categories provide interfaces
for expert-specified objectives, data, procedures, or resource constraints.
Verification combines reference equivalence, conformance tests, bounds on
scientific quantities, and re-execution as appropriate to the task.
Acceleration adds a resource criterion after correctness; reproduction
grades the outputs of a runnable procedure. This division reuses mechanical
authoring operations without transferring scientific judgment to them.

%% file: check_design_method.tex
\subsection{From official tests to scientific checks}
\label{sec:checks}

Once a module is approved, its upstream unit tests, regression tests, and
shipped example problems are surveyed together. Each case is traced to the
module responsibility, run or marked as unmeasured, and either retained as a
check, explicitly excluded with a reason, or identified as a gap. The unit of
reward is a \emph{check}: fixed inputs and graded outputs
paired with a \emph{pass policy}. An example remains an official test even
without a shipped reference: the pinned build supplies the comparison and
the example's physics (a published value, convergence, or a conserved
quantity) anchors it. A check with no official source is \emph{custom} and
requires curator agreement.

The pass policy has two forms. A \emph{pointwise} policy
compares every graded value,
$|c-r|\le a + \rho\,|r|$; exact equality is the special case
$a=\rho=0$. Pointwise is preferred whenever a bound contains the measured
sensitivity over a scientifically meaningful window while still rejecting a
real fault. It grades physical observables, never bookkeeping: an unordered
collection is aligned by an identity carried in the output before comparison
(a particle's properties follow its particle ID rather than its array slot),
and the same alignment covers its associated arrays. Storage order, adaptive
step counts, timings, rank or chunk layout, random draws, and eigenvector
phase are not observables. An \emph{invariants} policy compares quantities
such as moments, distributions, conserved values, or integral norms when no
pointwise bound can contain the run-to-run variation and reject a fault. This
includes random streams, rapidly diverging flows, sampling statistics, and
discrete outputs. The graded window is shortened first when the physics still
survives that choice; otherwise invariants are used from the start.

Nominal and variant initial conditions make this choice measurable. The
variant perturbs the smallest sufficient set of active inputs so that the
graded outputs reveal numerical sensitivity; it is calibration evidence, not
a physics-isolation experiment or an automatic tolerance rule. Where a build
permits it, an optional \emph{altbuild} runs the nominal input on another
legitimate build of the same pinned source and records the observed
cross-build separation. A check without such a build has no altbuild floor;
even a measured floor or nominal--variant spread is finite evidence, not a
universal guarantee. Self-validation runs nominal and variant independently,
requires each to reproduce the reference and attain full reward, and records
their observations separately. The curator then finalizes the policy,
tolerance, window, and variant by reading the source and its numerical
mechanisms. Each check carries a
plain-language \emph{warrant}: it identifies the observable, explains which
scientifically relevant bias the bound distinguishes, and explains why a
valid implementation can satisfy it. Implementation-level calibration details
are given in Appendix~\ref{app:task-construction-protocols}. The curator and
domain expert then review the package over fresh rounds, reproducing and
revising checks when needed; revised contracts require fresh evidence before
acceptance.

The check suite is fixed before downstream task authoring. Task statements
can therefore vary the requested work while every check remains part of the
acceptance contract.

%% file: sciencelobe.tex
\subsection{Manufacturing and validating scientific tasks}
\label{sec:manufacture}
\begin{figure}[!htbp]
  \centering
  \includegraphics[width=\linewidth]{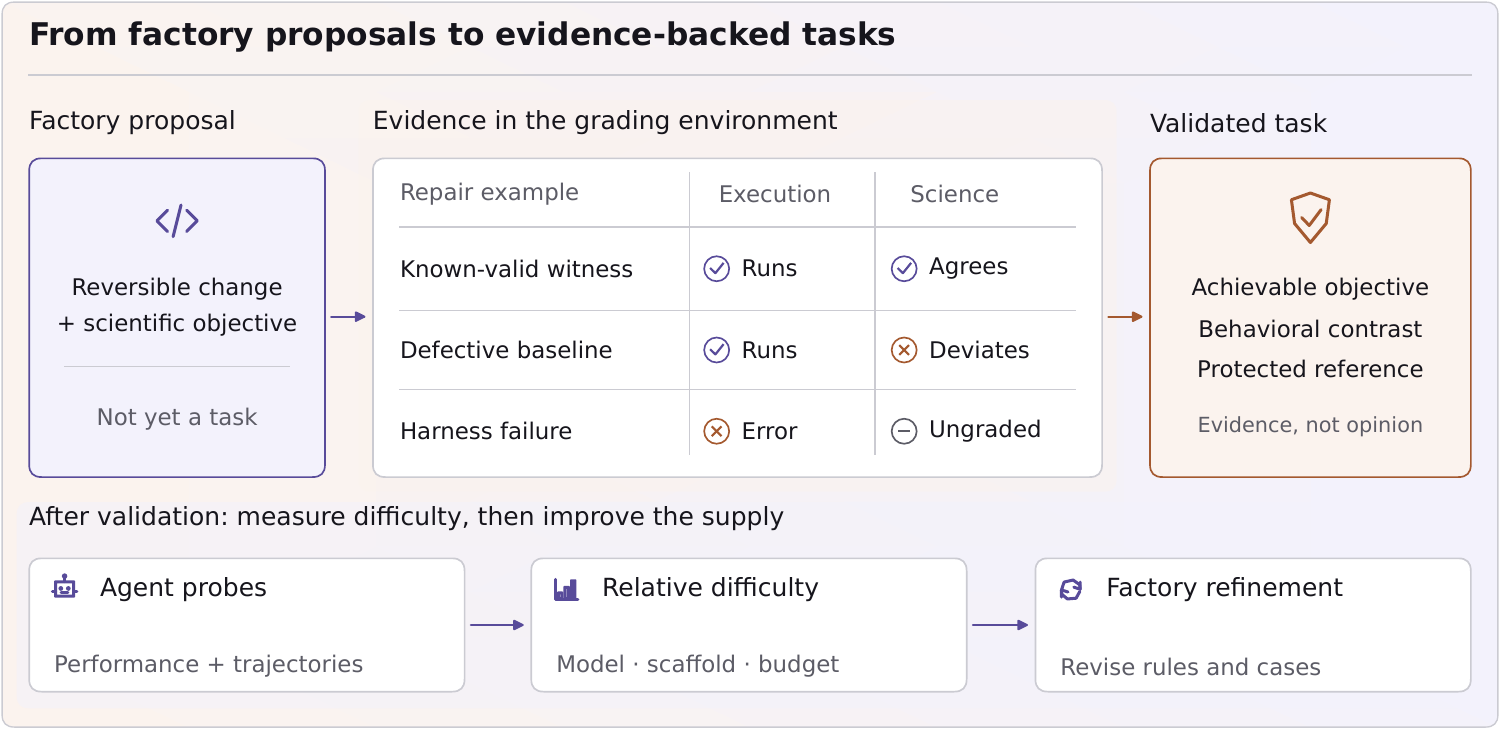}
  \caption{\textbf{Scientific task validity requires executable evidence.}
  For injected repairs, the known-valid witness and defective baseline both
  run, but only the witness satisfies the scientific checks. Harness failures
  remain ungraded.
  Design probes guide proposals; validated tasks receive model-relative
  difficulty measurements that inform factory refinement.}
  \label{fig:pipeline}
\end{figure}

Factory proposals become tasks only after executable evidence establishes that
they are observable, solvable, and trustworthy (Figure~\ref{fig:pipeline}).
AI authors can propose repairs, implementations, accelerations, reproductions,
or other transformations, but a proposal is not evidence of validity.

\paragraph{Propose broadly; establish validity by execution.}
\label{sec:validity}
Factory rules turn AI-proposed transformations into candidates, including
syntactic mutations~\citep{just2014mutants, just2014defects4j}, cross-component
changes, and non-repair objectives.
Each candidate is tested in the final grading environment: its objective must
be attainable and yield an informative score. For an injected repair, a
known-valid witness must pass, the unfixed baseline must leave headroom, and
the change must produce a check failure removed by the reference repair.
Where applicable, compilation, native execution, and coverage precede
admission; family-specific closed-book and visibility probes provide design
evidence rather than universal validity criteria. Silent mutations may remain
explicit controls, whereas ambiguous specifications, unreachable branches,
and harness failures are invalid or ungraded.

\paragraph{Trustworthy packaging and useful progress.}
\label{sec:reward}
Scientific checks expose partial accomplishment; repair scores normalize it
against the defective starting point:
\begin{equation}
 r_{\mathrm{repair}}=\max\!\left(0,\frac{r-f}{1-f}\right),
 \qquad 0\leq f<1,
 \label{eq:repair-reward}
\end{equation}
where $r$ measures scientific agreement and $f$ is the unfixed build's score.
Packaging also screens answer leakage and records execution-integrity failures;
the gate and self-test protocol is part of task construction. For objectives
other than repair, evidence must show an attainable target, observable outcome,
and informative credit for alternatives. The task record separates graded
outcomes from execution and infrastructure failures and preserves the failure
reason for audit.

\label{sec:floorgates}

\paragraph{Policy- and budget-relative difficulty.}
\label{sec:difficulty}
Design aims for clear specifications, substantial reasoning depth, legitimate
alternatives, and resistance to leakage. Scientific-check calibration and
probe-campaign accounting are summarized in
Appendix~\ref{app:task-construction-protocols}.

After validity, task-family admission labels remain distinct from difficulty
measured for a named model or panel, scaffold, budget, and date. Trace audits
distinguish solver failure and budget exhaustion from infrastructure faults,
ambiguous specifications, and reward misalignment; each model diagnosis needs
supporting evidence, and no judge can override deterministic anchors. Outcomes
are reported in \S\ref{sec:experiments} and feed back into factory rules, cases, and
budgets. Design labels remain design intent, not retrospective claims about
measured hardness.

%% file: posttraining.tex
\subsection{Connecting scientific interaction to model learning}
\label{sec:posttraining}
\label{sec:sciencelobe}

\begin{figure}[!htbp]
  \centering
  \includegraphics[width=\linewidth]{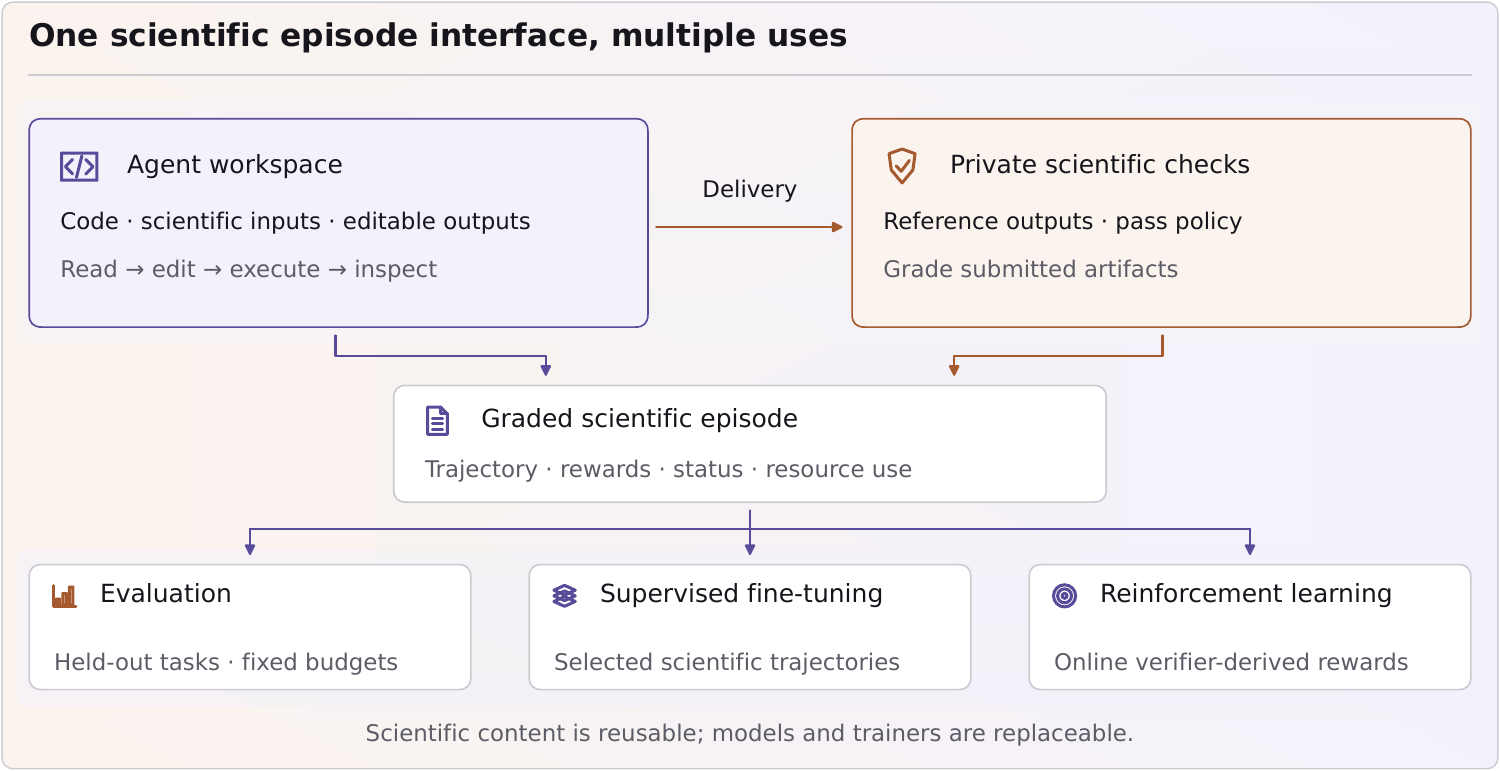}
  \caption{\textbf{A shared scientific episode interface supports evaluation and learning.}
  Agent workspaces submit artifacts to private scientific checks. The episode
  records trajectories, rewards, execution status, and resource use.
  Evaluation, SFT, and RL consume this experience through separate interfaces;
  scientific content remains reusable across models and trainers.}
  \label{fig:learning}
\end{figure}

A validated task exposes a common episode interface
(Figure~\ref{fig:learning}). The agent receives an editable workspace,
inspects code and scientific inputs, makes changes, runs experiments,
and submits artifacts to a private verifier. The harness records actions,
observations, check rewards, execution status, and resource use,
distinguishing scientific disagreement, incomplete delivery, and
infrastructure failure.

\paragraph{Evaluation, offline supervision, and online learning.}
\label{sec:evaluation}
Evaluation holds scientific tasks and interaction budgets fixed while
measuring success, partial progress, and resource use.
SFT consumers select trajectories and supervise agent actions, including
tool calls; RL consumers connect policy-controlled rollouts to verifier
rewards. Tasks can be selected by environment, domain, family, or measured
difficulty for held-out evaluation and training curricula.
Models and trainers can change without rebuilding the scientific content
or its acceptance criteria.

\paragraph{Returning capability to science.}
\label{sec:maintenance}
\label{sec:discovery}
The environments also support maintenance, acceleration, reproduction,
tool integration, and discovery-oriented work. Interactions supply
trajectories and proposals for new objectives; proposed tasks or checks
re-enter scientific validation before joining the registry.
The same interface thus supports learning from scientific work and
applying the resulting agents to further work.

%% file: experiments.tex
\section{Experiments}
\label{sec:experiments}\label{sec:results}

We evaluate scientific task execution in ScienceIDE and the benefits of
learning from scientific interactions. Agent comparisons use the
ScienceIDE-Hard subset; separate SFT and RL studies measure
public-benchmark transfer and learning on held-out scientific tasks.

\subsection{Experimental Setup}
\label{sec:setup}
\label{sec:scaling}
\label{sec:collection-diagnostics}

\paragraph{Scientific experience supply.}
ScienceIDE provides \RegistryEnvs{} environments from
\RegistrySources{} scientific codebases and \RegistryTasks{} manufactured
tasks (Figure~\ref{fig:registry}), of which repair and implementation supply
nearly all (2,515 and 295) and acceleration two so far; each environment
designates the workload an acceleration task must speed up. A companion collection provides 1,076 executable checks of numerical
outputs and physical invariants. Inventory, calibration, and source
details are in Appendices~\ref{app:experience-inventory}
and \ref{app:envsources}.

\paragraph{ScienceIDE-Hard.}
For public evaluation, we designate a validated subset of 85 hard tasks
from 18 environments derived from PLUTO, Athena++, MITgcm, LAPS,
and PHANTOM, covering astrophysical flows, plasma physics,
ocean and atmospheric modeling, and particle simulations.
Its 52 repair and 33 implementation tasks require agents to correct
defective code or reconstruct missing functionality, then deliver the
required scientific outputs. Selection combines reference-solver
difficulty probing with executable validation of the reference solution
and defective baseline (Appendix~\ref{app:gateaudit}).
Success requires agreement with private scientific references under
the task's acceptance criteria, not merely successful execution.

\paragraph{Agents and execution conditions.}
We compare fifteen models from eight providers through Codex, Claude Code,
or Gemini CLI. Each receives the same task-specific container and
instructions, no additional localization hints, and a one-hour episode
budget. Results therefore compare model--harness systems under a common task
interface. Figure~\ref{fig:eval-leaderboard} identifies the models and harnesses.

\paragraph{Metrics and learning comparisons.}
The primary metric is strict scientific success
($r_{\mathrm{repair}}=1$); incomplete or budget-exhausted deliveries
are unsuccessful. Tasks are weighted equally, within-task repeats are
averaged, and intervals describe execution variability on the fixed
test set. Estimators, effort accounting, and repeat coverage are in
Appendix~\ref{app:agent-evaluation-protocol}; learning-specific datasets
and splits are given with the SFT and RL experiments.

%% file: results.tex
\subsection{Scientific Agent Evaluation}
\label{sec:evalresults}

\subsubsection{Overall scientific task performance}
\label{sec:overall-leaderboard}

\begin{figure}[!htbp]
  \centering
  \includegraphics[width=\linewidth]{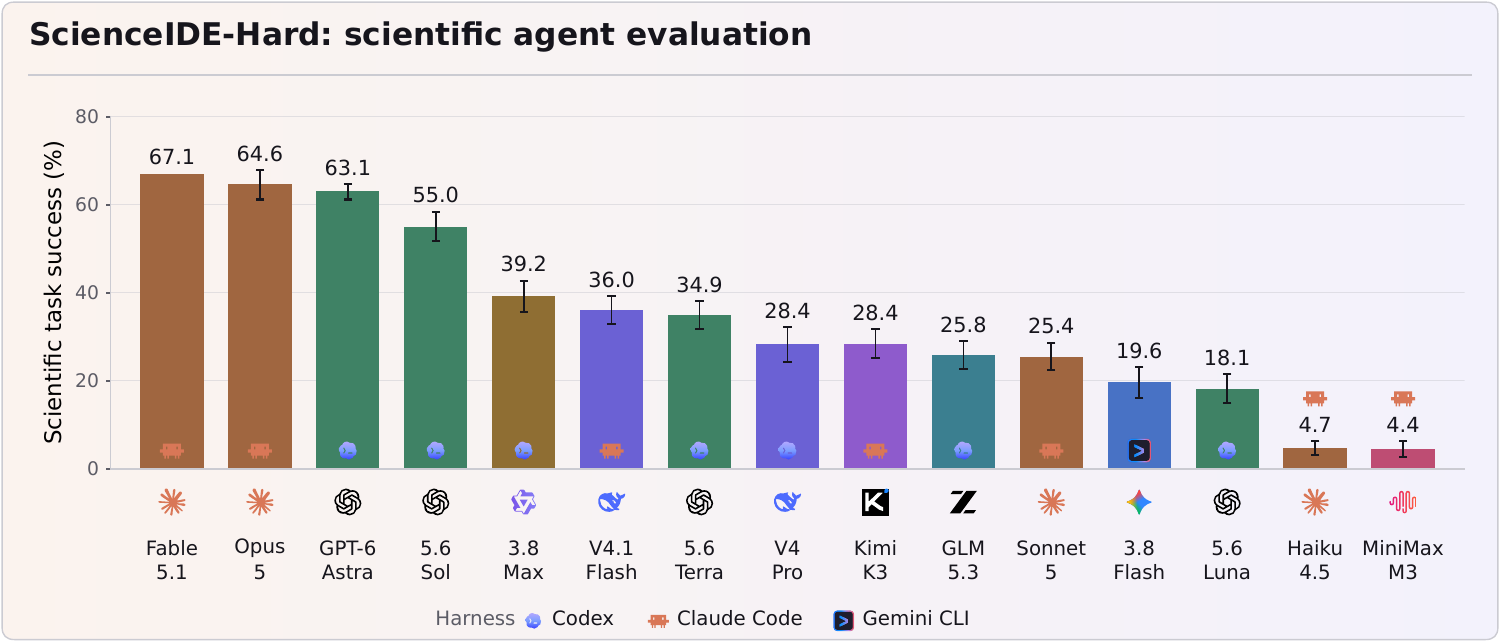}
  \caption{\textbf{ScienceIDE-Hard: scientific agent evaluation.}
  Strict success on the same 85 scientific repair and implementation tasks
  under a one-hour budget. Icons identify model providers and execution
  harnesses. Error bars show 95\% within-task repeat-bootstrap intervals;
  Fable has a single-attempt estimate. Measurement and repeat-coverage details are in
  Appendix~\ref{app:agent-evaluation-protocol}.}
  \label{fig:eval-leaderboard}
\end{figure}

Figure~\ref{fig:eval-leaderboard} compares all fifteen agents on the
85-task ScienceIDE-Hard subset under the protocol in \S\ref{sec:setup}, chosen
to probe multi-site edits and convention-faithful reconstruction in codebases
of $10^4$ to $10^5$ lines.

Fable 5.1 has the highest observed success rate at 67.1\%, followed by
Opus 5 at 64.6\% and Astra at 63.1\%. Sol reaches 55.0\%, while the remaining
eleven agents score below 40\%. Thus even the leading agents leave
roughly one third of these scientific tasks unsolved under the stated
budget. The top point estimates should not be read as a statistically
established ordering: Fable is singly measured, and the repeat intervals
of Opus and Astra overlap.

\subsubsection{Budget and resource efficiency}
\label{sec:conditional-capabilities}

\input{results_visuals}

Failure behavior, scientific specialization, and trajectory-level
error analysis are presented in Appendix~\ref{app:agent-diagnostics}.

\subsection{Learning from Scientific Trajectories}
\label{sec:sftresults}

\input{SFT}
\FloatBarrier

\subsection{Learning through Scientific Interaction}
\label{sec:rlresults}

\input{RL}
\FloatBarrier

%% file: results_visuals.tex
\begin{figure}[!htbp]
  \centering
  \includegraphics[width=\linewidth]{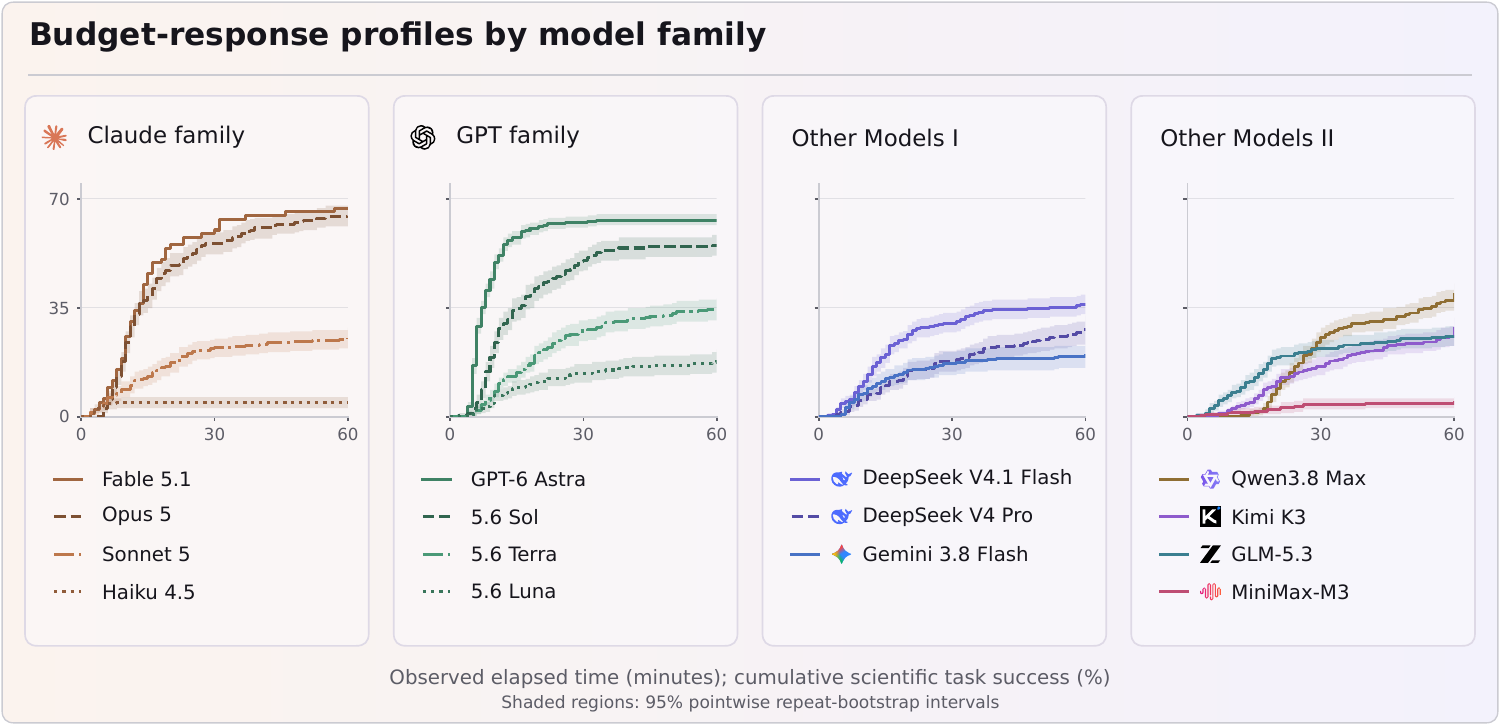}
  \caption{\textbf{Budget changes the observed ordering.}
  Cumulative scientific task success over the recorded one-hour episodes
  for all fifteen agents. Shading denotes pointwise 95\% within-task
  repeat-bootstrap intervals where available. These are retrospective completion profiles, not reruns under
  shorter assigned budgets.}
  \label{fig:budget}
\end{figure}

\paragraph{Agents differ in how quickly they solve scientific tasks.}
At ten minutes, Astra reaches 49.6\% success against Fable's 25.9\%;
Fable overtakes it at approximately 31 minutes
(Figure~\ref{fig:budget}). Between 20 and 60 minutes, Astra gains only
2.0 percentage points, compared with 11.8 for Fable, 16.0 for Opus, and
30.2 for Qwen3.8 Max. Similar final scores can therefore conceal substantially
different time requirements.

\begin{figure}[!htbp]
  \centering
  \includegraphics[width=\linewidth]{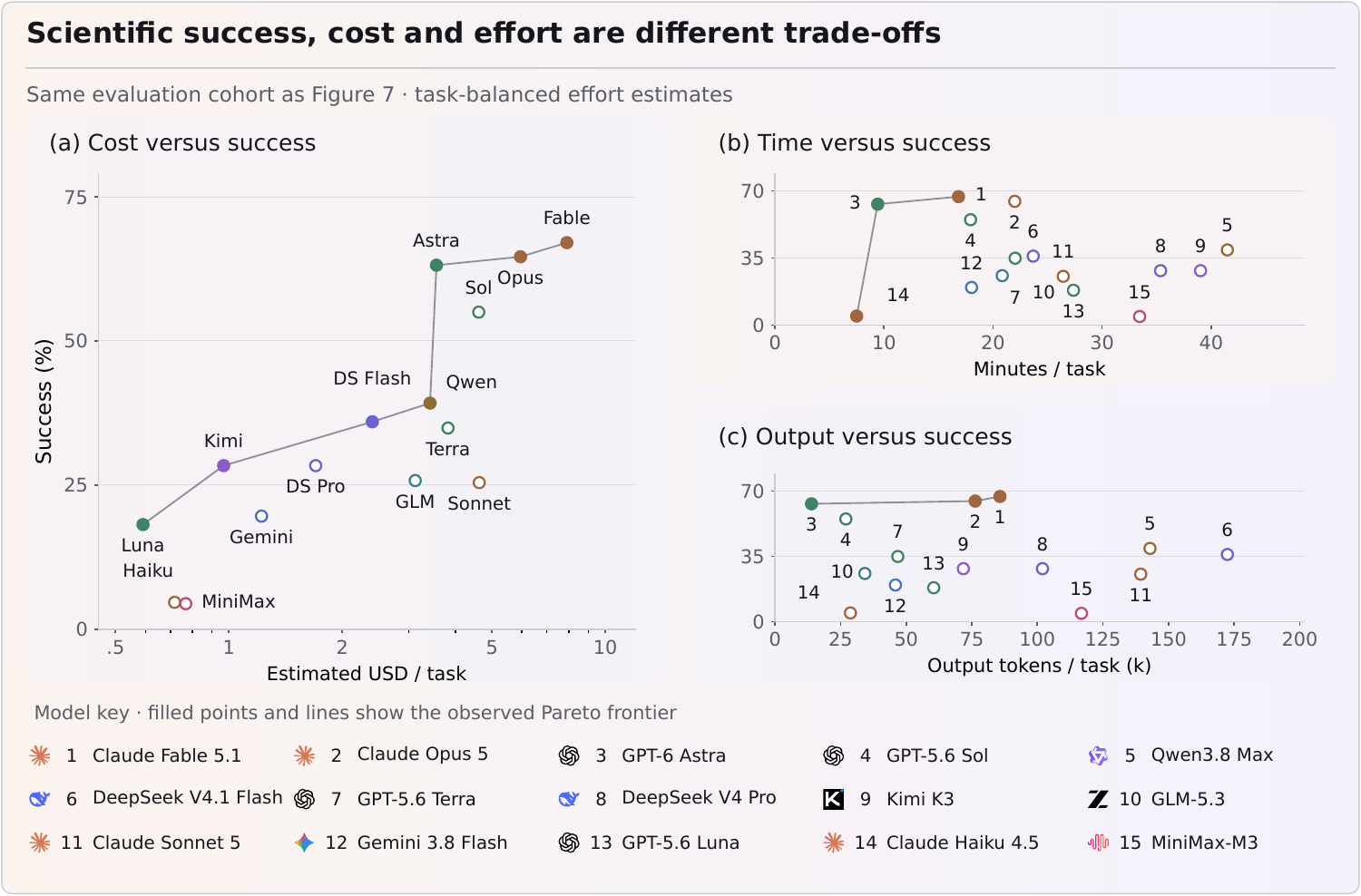}
  \caption{\textbf{Outcome--cost profiles separate accuracy from expenditure.}
  Success rates match the leaderboard for all fifteen agents. Effort averages
  available telemetry within each task from the same selected measurements.
  Filled markers and connecting lines show observed Pareto frontiers. Costs use recorded
  harness estimates or archived token rates rather than current prices.}
  \label{fig:efficiency}
\end{figure}

\paragraph{More expenditure does not guarantee greater success.}
On the leaderboard cohort, Fable achieves 67.1\% success at an estimated
\$7.90 per task, while Astra achieves 63.1\% at \$3.56
(Figure~\ref{fig:efficiency}). Astra averages 9.4 minutes and 13.9k output
tokens per task, compared with Fable's 16.8 minutes and 85.7k tokens.
DeepSeek V4.1 Flash produces 172.4k output tokens per task for 36.0\%
success. Across the fifteen profiles, the descriptive Spearman
correlations of success with runtime and output volume are $-0.22$ and
$0.01$. Resource use and scientific correctness are thus distinct
dimensions of performance.

%% file: SFT.tex

Scientific interaction trajectories record how models inspect code,
use tools, and respond to execution feedback. We ask whether learning
from these demonstrations improves scientific-code repair and transfers
to public benchmarks. We fine-tune Qwen3.5-4B, Qwen3.5-9B, and
Qwen2.5-72B-Instruct on verified ScienceIDE demonstrations and compare
each final checkpoint with its initial model under matched settings.

\paragraph{Demonstrations and supervision.}
We collect demonstrations with GPT-5.6-sol and select trajectories
using the numerical-equivalence verifier. The training partition
contains 4,567 segments from 564 tasks, and the validation partition
contains 544 segments from 81 tasks. We retain the source partition
assignments; no task identifier appears in both partitions.

\paragraph{Training and validation.}
We train with ms-swift, applying LoRA to all linear layers for three
epochs, and evaluate the final checkpoints without selecting them on
public-benchmark performance. Example construction, hyperparameters,
and teacher-forced validation loss and token accuracy are reported in
Appendix~\ref{app:sft-training-details}. These diagnostics measure
prediction of the demonstrations; we assess scientific repair
separately using the numerical checks below.

\paragraph{Public-benchmark evaluation.}
We evaluate the initial and SFT models on public benchmarks covering
scientific knowledge, mathematical reasoning, code understanding,
repair, and generation. Within each model pair, both checkpoints
are evaluated on identical items. Scoring rules and generation settings
are given in Appendix~\ref{app:sft-training-details}; scores from
different metrics are reported separately, without an overall average.

\paragraph{Scientific-code repair evaluation.}
We evaluate repair tasks whose identifiers are excluded from both
training and validation. Each prompt supplies the task description,
localized source excerpts, and candidate routines. The model produces
one structured patch, which a fixed runner applies before building the
code and running the environment's original numerical checks. Initial
and SFT checkpoints receive identical inputs and use greedy decoding,
an 8,192-token context limit, and a 2,048-token response limit. For each
environment with complete paired evaluations, we report mean repair
reward $r_{\mathrm{repair}}\in[0,1]$, retaining partial credit. This protocol evaluates localized
repair on held-out tasks within the existing scientific codebases.

\begin{figure}[t]
\centering
\includegraphics[width=\linewidth]{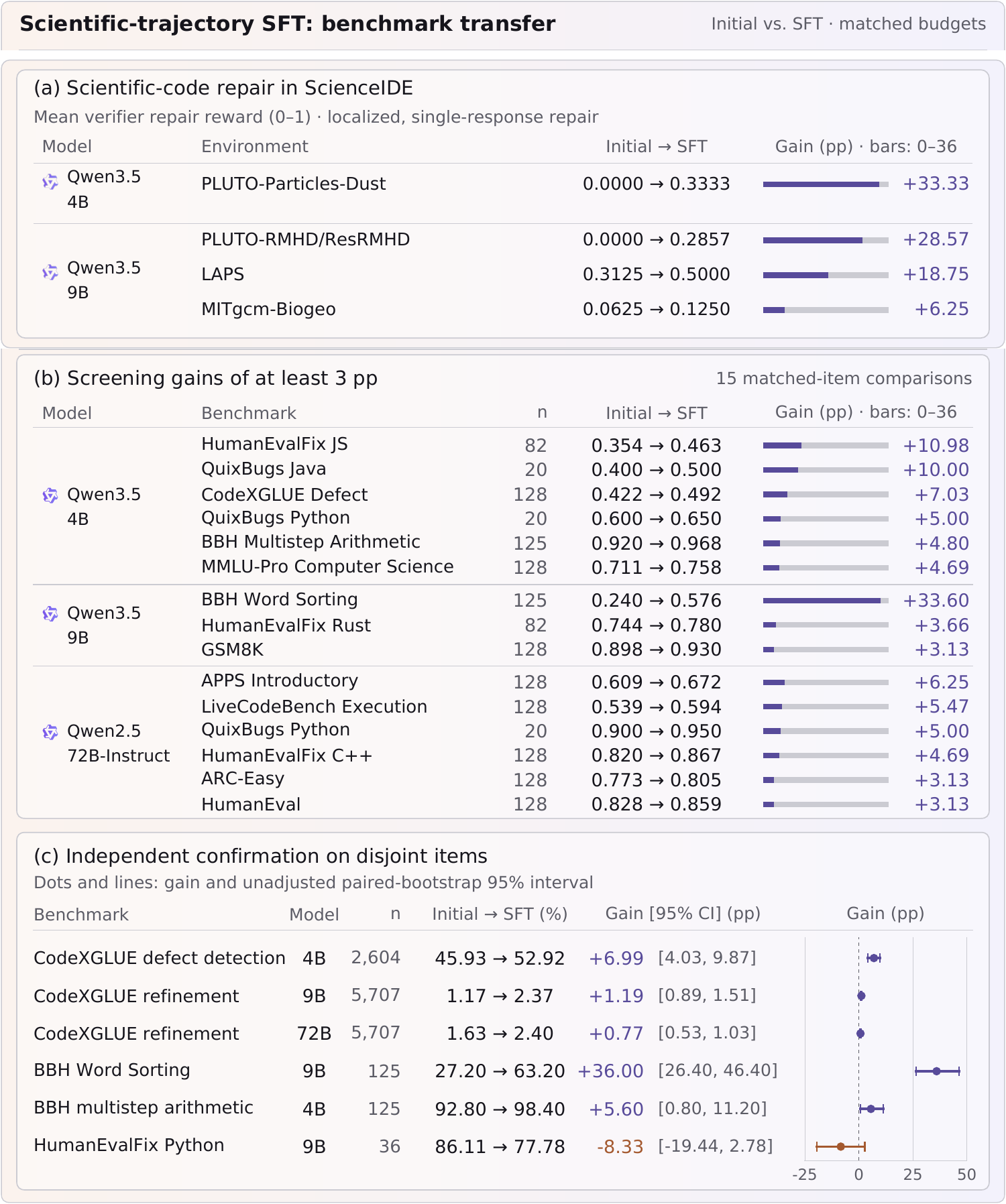}
\caption{\textbf{Benchmark improvements.}
(a) Mean verifier repair reward on a 0--1 scale for held-out ScienceIDE
tasks under matched single-response settings, grouped by model. Gains
are reported in percentage points (100 times the reward difference),
consistent with the gain columns in (b) and (c).
(b) The 15 model--benchmark comparisons with screening gains of at
least 3 percentage points, grouped by model. Rows give initial
$\rightarrow$ SFT scores on a 0--1 scale; gains are computed before rounding.
(c) Confirmation comparisons on disjoint items with unadjusted
paired-bootstrap 95\% intervals for the gain, including one decline
(Appendix~\ref{app:sft-training-details}).}
\label{fig:sft-benchmark-results}
\end{figure}

\paragraph{SFT improves repair reward across scientific environments.}
Figure~\ref{fig:sft-benchmark-results}(a) shows higher mean repair reward in
three scientific codebases. For 4B, mean reward on PLUTO-Particles-Dust
rises from 0.0000 to 0.3333 (+33.33 points). For 9B, it rises from 0.0000 to 0.2857 (+28.57 points) on
PLUTO-RMHD/ResRMHD, from 0.3125 to 0.5000 (+18.75 points) on LAPS, and from 0.0625 to
0.1250 (+6.25 points) on MITgcm-Biogeo. These gains are measured by the environments'
original numerical checks under the same repair protocol.


\paragraph{Gains span code, reasoning, and knowledge tasks.}
Figure~\ref{fig:sft-benchmark-results}(b) shows 15 model--benchmark
comparisons with improved scores. The code gains span repair,
execution prediction, defect detection, and generation. For 4B,
HumanEvalFix JavaScript improves by 10.98 percentage points, and
QuixBugs Java increases from 0.400 to 0.500 (+10.00 points).
CodeXGLUE defect detection increases from 0.422 to 0.492
(+7.03 points), although final accuracy remains below 0.5.

Smaller code gains appear for 9B and 72B, and the improvements extend
to reasoning and knowledge benchmarks. For 9B, BBH Word Sorting
increases from 0.240 to 0.576 (+33.60 points), alongside a gain on GSM8K.
Other improvements include 4B on BBH multistep arithmetic and MMLU-Pro
Computer Science, and 72B on ARC-Easy.
Appendix~\ref{app:sft-training-details} reports the detailed comparisons.

These results suggest that scientific interaction trajectories can
support both scientific specialization and broader capability
development. Beyond improved repair reward on held-out ScienceIDE
tasks, the observed gains extend to public benchmarks in code,
reasoning, and knowledge, providing evidence of positive transfer
from scientific experience to general capabilities. Although these
gains do not establish uniform improvement or frontier-level
performance, they provide an empirical basis for pursuing a joint
goal: leading scientific capabilities alongside competitive
general-purpose performance. Scientific experience thus offers a
promising training resource for advancing both.

%% file: RL.tex
Scientific environments also provide a reward signal that supervised data cannot:
a verifier scores an agent's repair by compiling the modified code and running
the underlying scientific simulation. We use this signal directly as the RL
reward and ask whether online optimization can improve Qwen3.5-4B beyond its
base checkpoint. We train on two environments, LAPS, a 3D pseudo-spectral
Hall-MHD solver \citep{Shi2024LAPS}, and MITgcm-biogeo, which models ocean
biogeochemistry and CFC transport. Their task banks contain 99 and 87 repair
tasks over three and nine graded scientific cases, respectively. Each repair is graded in $[0,1]$ with partial credit. The instructions identify the defect's file, line, and edit class, leaving the agent to determine what the repaired code should compute.

This setting makes RL substantially harder than conventional
tasks. A verifier verdict arrives only after the episode ends, so reward is
outcome-only over trajectories that may span tens of turns and tens of
thousands of tokens. Rollout and training run concurrently on disjoint GPUs, with token-level truncated importance sampling correcting the resulting one-step policy staleness (Appendix~\ref{app:rl-implementation}).

\begin{figure}[!t]
  \centering
  \includegraphics[width=\linewidth]{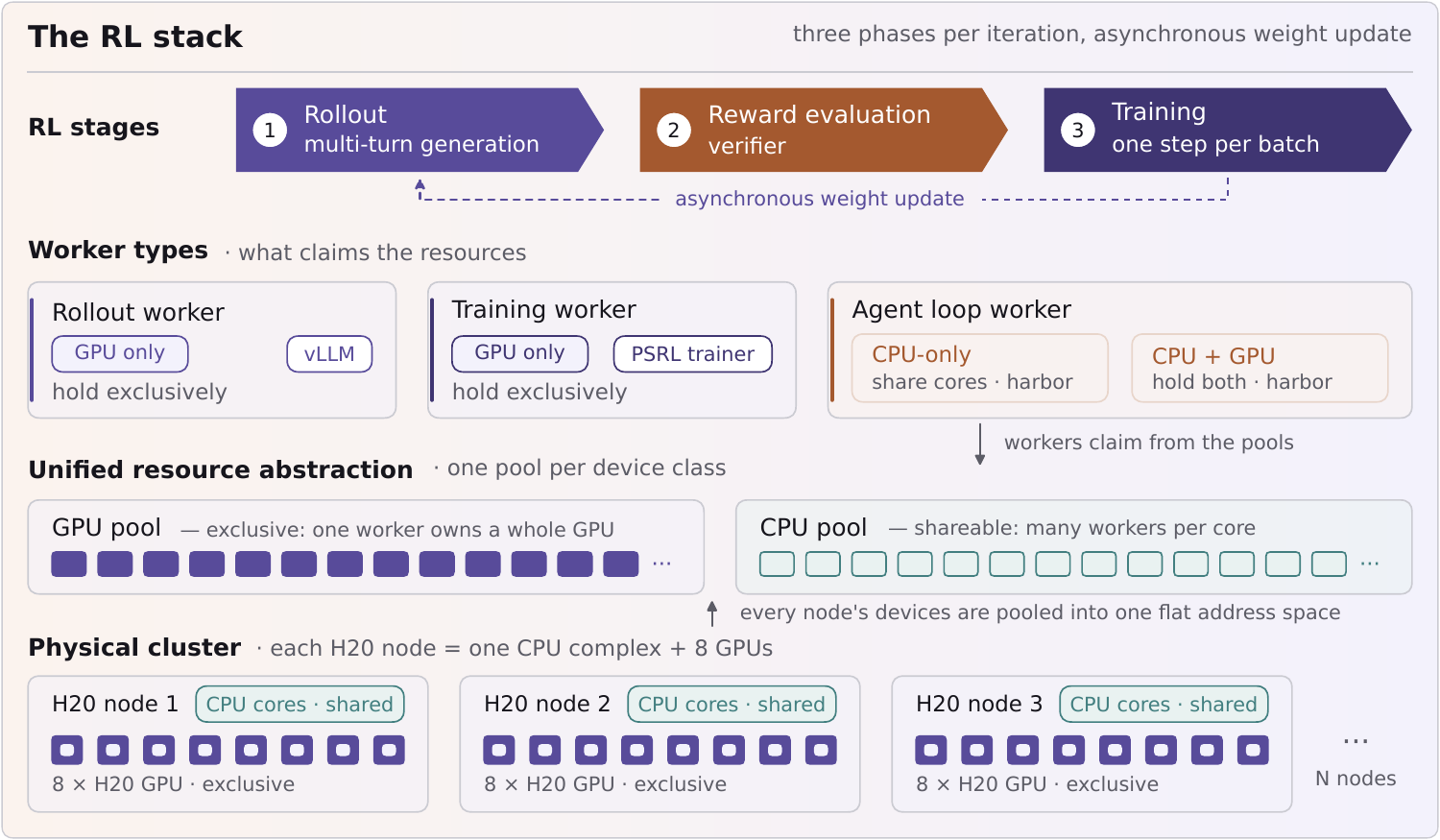}
  \caption{\textbf{The RL stack.} Three phases per iteration with
  asynchronous weight updates. Rollout workers (vLLM) and training workers
  (PSRL trainer) exclusively occupy GPUs, while agent-loop workers (harbor)
  that drive episode execution either share CPU cores or exclusively occupy
  both CPUs and GPUs.}
  \label{fig:rl_overview}
\end{figure}

\paragraph{Execution and training stack.}
Generation runs in vLLM \citep{kwon2023vllm} and optimization in PSRL
\citep{li2026staleflow, zhou2026harnessing}, our customization of the
veRL trainer \citep{sheng2024verl} that drives a black-box harness;
episodes execute in harbor through the same containers and verifiers as
\S\ref{sec:evalresults}, so training consumes the verifier's native
reward rather than a learned proxy. Figure~\ref{fig:rl_overview} shows
the three phases of an iteration, multi-turn rollout, verifier reward
evaluation, and one optimizer step, with weights updated asynchronously.
Rollout and training workers each own whole GPUs, while the agent-loop
workers that drive episode execution share CPU cores or claim both CPUs and
GPUs, and every node's devices join one pool per device class.
Episodes are expensive and uneven in length, since one may rebuild a
Fortran solver and run a full simulation, so a synchronous loop would wait
for the slowest episode of every batch. Giving rollout and training
disjoint GPUs lets generation continue against the current weights while
the optimizer runs; on a representative step the update took 2,069 of
3,210 seconds and rollout 751 seconds. Agent-loop workers are light CPU processes, so many episodes execute concurrently on shared cores (Appendix~\ref{app:rl-implementation}).

\paragraph{Objective.}
For a prompt group of $G=8$ trajectories with rewards $R_i$, the
group-relative advantage is
\begin{equation}
  \hat{A}_i = R_i - \frac{1}{G}\sum_{j=1}^{G}R_j ,
  \label{eq:rl-adv}
\end{equation}
which omits the division by the group standard deviation of the original
formulation, following Dr.~GRPO \citep{drgrpo}, because near-bimodal
group rewards would otherwise inflate isolated successes
(Appendix~\ref{app:rl-implementation}). Tokens are then updated with a PPO-style
clipped ratio whose bound is asymmetric ($\varepsilon_{\mathrm{low}}=0.2$,
$\varepsilon_{\mathrm{high}}=0.3$), so that rarely sampled repair actions
retain room to gain probability mass \citep{dapo}; the loss is aggregated
by token mean; Eq.~\eqref{eq:rl-clip} and the remaining settings, unchanged across both runs, are in Appendix~\ref{app:rl-implementation}.


\paragraph{Budget cutoffs are not failed repairs: a long-horizon challenge in science RL.}
In long-horizon scientific tasks, an episode can end because the repair
finishes, the turn cap is exhausted, or the response budget is exhausted.
Only the first indicates whether the repair succeeded, yet outcome-only reward scores all three alike. When a budget-truncated episode receives zero reward in a group with positive mean reward, Eq.~\eqref{eq:rl-adv} gives it $\hat{A}_i<0$, suppressing its generated tokens under the unmasked objective. If all group rewards are zero, its advantage is zero. With token-mean aggregation, this penalty is especially strong for the very
trajectories that consume the most tokens. The objective consequently creates
a shortcut: the policy can improve either by solving the scientific task or by
simply making its trajectories shorter.

The effect is especially harmful because the longest trajectories are often
the ones attempting the hardest repairs. Penalizing them for exhausting a
budget therefore discourages the policy from exploring solutions that require
extended interaction with the scientific environment. We observe this failure
mode directly in an unmasked run. Reward initially improves but then collapses
below its starting point, while tokens per turn fall by more than a factor of
three. Shorter turns cause more trajectories to exhaust the turn cap, which
increases truncation and reinforces the same pathological signal. The model
can thus enter a self-reinforcing cycle in which shortening the trajectory
looks preferable to making progress on the repair.

The fix is to separate the role of a truncated trajectory as a reward
observation from its role in policy optimization. A harness cutoff says
nothing about whether the tokens generated before the cutoff were useful, so
those tokens should not receive a policy gradient. At the same time, the
trajectory's reward remains informative for the group baseline. We therefore
keep budget-truncated trajectories in the group baseline and mask them out of
the loss (Eq.~\eqref{eq:rl-mask} in Appendix~\ref{app:rl-implementation}).
Its reward still enters Eq.~\eqref{eq:rl-adv}; completed episodes receive a positive advantage when their reward exceeds the group mean. Completion alone does not guarantee a positive advantage. By the same reasoning, we switch off DAPO's reward-side length penalty.
Length here reflects the difficulty of localizing and repairing a defect, not a quantity to penalize (paired comparison and penalty audit in Appendix~\ref{app:rl-implementation}).
\FloatBarrier

\begin{figure}[!htbp]
\centering
\includegraphics[width=\linewidth]{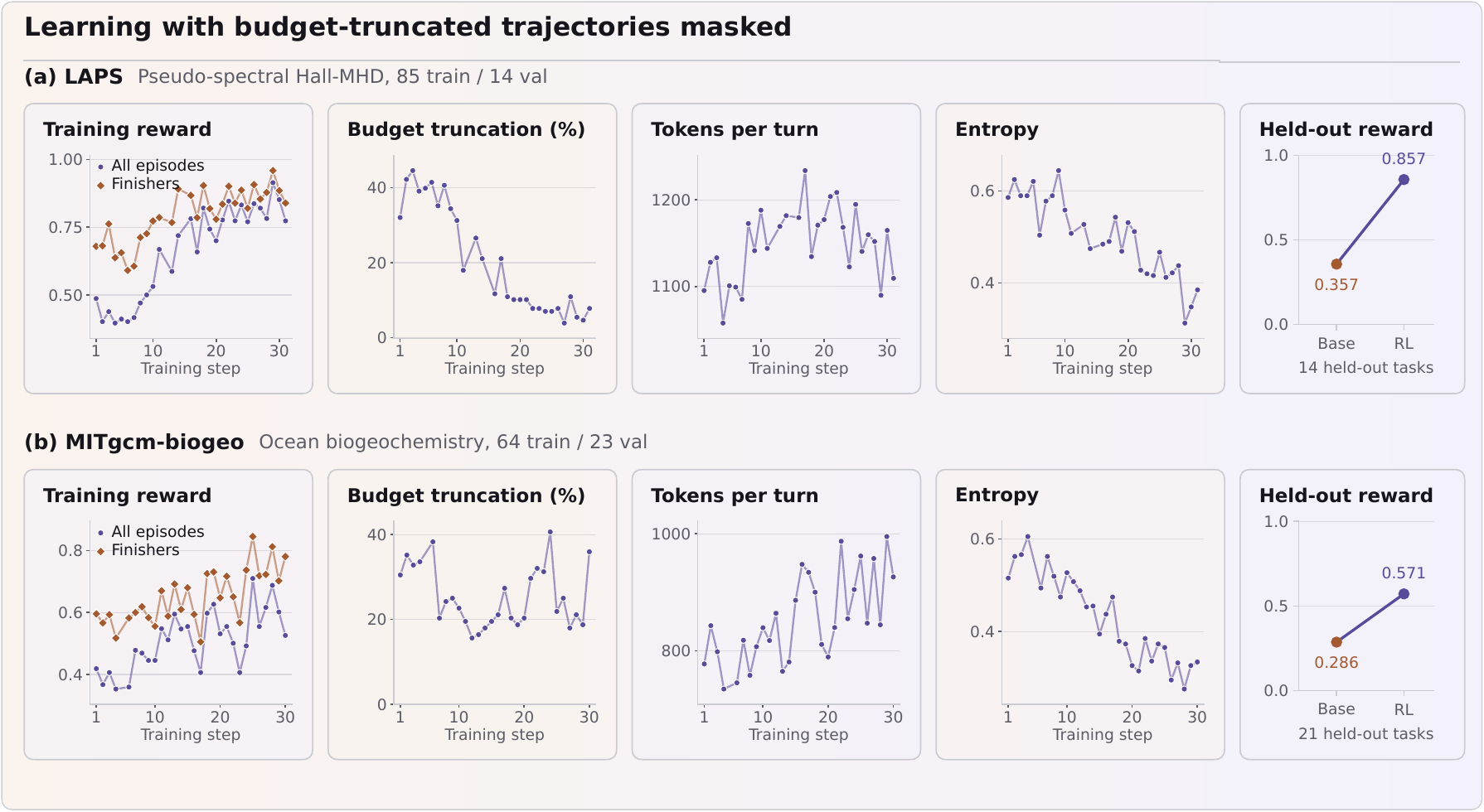}
\caption{\textbf{Training dynamics with budget-truncated trajectories masked,
and held-out reward.} Every recorded step of both runs; headers give source bank splits (the logged MITgcm-biogeo validation set has 21 of its 23 tasks, Appendix~\ref{app:rl-implementation}). Right column: held-out reward of base Qwen3.5-4B and its step-30 checkpoint on 14 LAPS and 21 MITgcm-biogeo tasks under matched harnesses and hints; values are mean shaped verifier rewards.}
\label{fig:rl-training}
\end{figure}

\paragraph{Online verifier feedback substantially improves held-out scientific reward.}
With the truncation mask in place, held-out reward rises in both environments (Figure~\ref{fig:rl-training}, right column): after 30 training steps, LAPS from $0.357$ to $0.857$ ($2.4\times$) and MITgcm-biogeo from $0.286$ to $0.571$ ($2.0\times$) over the base Qwen3.5-4B checkpoint.

The training dynamics show concurrent gains in reward and reductions in
truncation (Figure~\ref{fig:rl-training}). Comparing the first and last five
recorded steps, LAPS mean training reward increases from
$0.427$ to $0.828$, while mean reward among completed episodes rises from
$0.683$ to $0.883$. At the same time, budget truncation drops from $39.5\%$ to
$6.6\%$. MITgcm-biogeo shows the same trend, with reward increasing from
$0.381$ to $0.597$, mean reward among completed episodes increasing from $0.571$ to $0.747$,
and truncation decreasing from $34.1\%$ to $23.8\%$. The policy increasingly
converts long-running attempts that previously died at a budget cutoff into
episodes that reach a verifier verdict and receive meaningful feedback.

Truncation falls without shorter responses: mean tokens per turn increase
from 1103 to 1135 on LAPS and from 780 to 914 on MITgcm-biogeo.

Both environments show higher training and held-out reward with less
truncation, supporting verifier-guided scientific repair across simulators
under the same long-horizon training procedure.

%% file: relatedwork.tex

\section{Related Work}
\label{sec:related}

AI and science interact in two complementary directions
(Figure~\ref{fig:teaser}). \emph{AI for Science} uses models and agents for
scientific prediction, computation, and discovery; \emph{Science for AI}
draws on scientific ideas, data, and environments to develop AI capabilities.
A scientific result does not necessarily improve the underlying model,
just as a high benchmark score does not establish a reusable training
environment. Connecting these directions requires more than successful task
execution: scientific workflows must expose actions and outcomes that can
be validated, reused, and converted into learning experience.

\subsection{AI for Science: models, agents, and evaluation}

\paragraph{Specialist scientific models.}
Scientific models span protein structure prediction \citep{alphafold},
PDE learning through neural operators and physics-informed networks
\citep{fourier-neural-operator, pinns}, weather forecasting
\citep{graphcast, pangu-weather, fourcastnet}, materials generation
\citep{mattergen}, symbolic regression \citep{ai-feynman}, and
scientific language and reasoning \citep{galactica, minerva}.
These approaches develop domain capabilities through specialized models,
objectives, or corpora; agent systems additionally organize model actions
into scientific workflows.

\paragraph{Scientific agents and discovery.}
Reported advances cover mathematics, physics, and biology. Examples include
Claude's computer-checked Fermat formalization \citep{claude-flt}; GPT-5.2's gluon-amplitude
conjecture, subsequently proved, and GPT-5's contributions to an
Erd\H{o}s problem and an experimentally confirmed immune-cell mechanism
\citep{gpt52-physics, gpt5-science}. AlphaEvolve extends FunSearch's
program-search approach \citep{funsearch}, reporting an improvement over
Strassen's $4\times4$ complex-valued matrix-multiplication algorithm
and advances in Ramsey bounds \citep{alphaevolve,nagda2026ramsey}.
Biological examples include co-scientist drug-repurposing hypotheses
\citep{ai-co-scientist}, Robin's ripasudil proposal for macular degeneration
tested in ARPE-19 retinal pigment epithelial cells \citep{robin}, and
discoveries reported
by Kosmos \citep{kosmos}. Broader research frameworks pursue autonomy
through tool use, search, orchestration, and collaboration
\citep{ai-scientist, ai-scientist-v2, agent-laboratory, curie, chemcrow,
coscientist2023, sciagents, yang2026discovery, xue2026sciencebuddy}. These results demonstrate several forms of scientific validation, from
formal checking to experimental tests. ScienceIDE focuses on making such
execution and verification interfaces reusable across tasks and models.

\paragraph{Measuring scientific ability.}
Evaluation settings differ in task source and verification.
Expert-authored or publication-derived suites include SciCode
(80 problems) \citep{tian2024scicode}, ScienceAgentBench (102 tasks)
\citep{chen2024scienceagentbench}, LAB-Bench \citep{lab-bench},
the systems-biology SciGym \citep{duan2025scigym}, and DiscoveryBench
\citep{discoverybench}. Research-level evaluations span ML engineering
\citep{mle-bench}, expert-referenced R\&D \citep{re-bench},
frontier mathematics \citep{frontiermath}, and broad expert questions
\citep{hle}; CORE-Bench and PaperBench instead grade the reproduction
of existing results \citep{siegel2024corebench, paperbench}.
Closest to our software substrate, AInsteinBench sources maintainer
pull requests from six scientific repositories, with expert review,
annotated difficulty, and a requirement for well-structured test suites
\citep{duston2025ainsteinbench}. SWE-bench Science covers 119 tasks from
98 repositories and emphasizes how software tests can under-specify
scientific contracts \citep{xu2026swebenchscience}.
Expert authoring cost, saturation, and public-answer contamination
motivate reusable task-generation and verification mechanisms alongside
these evaluation resources.

\subsection{Science for AI: ideas, data, and learning environments}

\paragraph{Scientific foundations and supervision.}
Science contributes theoretical ideas, training data, and evaluation
problems. Statistical mechanics informs energy-based models, while
nonequilibrium thermodynamics inspired diffusion models
\citep{sohl-dickstein}. Structural, atmospheric, and materials databases
support the specialist models above, and scientific corpora support
language-model pretraining. Graduate-level science and research-level
mathematics provide reasoning evaluations such as GPQA, FrontierMath,
and HLE \citep{gpqa, frontiermath, hle}. Evaluation on these problems
does not itself establish that they supplied training gradients.
Interactive learning adds another channel: an agent acts, observes
consequences, and receives feedback from an executable environment.

\paragraph{Executable environments and the coding precedent.}
The learning stack separates environments, action interfaces, and
optimization. Game, procedural, embodied, web, and multi-agent settings
provide interaction \citep{ale, openai-gym, gymnasium, procgen, nethack-le,
crafter, minedojo, textworld, alfworld, webarena, pettingzoo, openspiel};
algorithm libraries support training \citep{raffin2021sb3,
huang2022cleanrl, torchrl}. Language-model agent suites
\citep{reasoning-gym, textarena, lmrl-gym, tau-bench, agentbench, gaia}
complement training and inference systems \citep{openrlhf, roll,
liu2024oat, tan2025rl2, cao2025skyrl, sheng2024verl, kwon2023vllm,
rasley2020deepspeed} and verifier-based supervision
\citep{lets-verify-step-by-step, star, beyond-human-data-rest}.
Tool-use scaffolds enable reading, editing, and execution
\citep{react, reflexion, codeact, swe-agent, openhands, voyager}.
Within coding, SWE-bench established repository evaluation
\citep{swe-bench}; SWE-Gym provides 2,438 real tasks with runtimes
and tests \citep{pan2024swegym}, R2E-Gym supplies 8.7k procedural
tasks with hybrid verifiers \citep{jain2025r2egym}, debug-gym exposes
interactive debugging \citep{yuan2025debuggym}, and SWE-smith
synthesizes 50k task instances \citep{yang2025swesmith}.
SWE-RL learns from software evolution \citep{swe-rl}, alongside broader
reasoning and agent training with verifiable feedback
\citep{deepseek-r1, kimi-k15, dapo, o1-system-card, openai2025gpt5,
anthropic2025claude4}. Formal mathematics provides a related example
through proof-checker feedback \citep{deepseek-prover}. Together these
directions give concrete mechanisms for learning from experience
\citep{silver2025experience}.

\paragraph{Scientific interaction environments.}
Existing scientific learning settings expose different kinds of action:
RL for tokamak control \citep{tokamak-control}, molecular generation
with task-specific rewards \citep{gflownets}, and physics-simulator
feedback for Olympiad reasoning \citep{physics-olympiad-rl}.
Solver-control gyms expose dynamical systems for control
\citep{lagemann2026hydrogym, bhan2024pdecontrolgym, zhang2023controlgym},
with safe-control environments studying constraints on action
\citep{ray2019safetygym}; GEM, MLGym, Aviary, and DiscoveryWorld
provide broader agent interaction settings
\citep{liu2025gem, mlgym, aviary, discoveryworld}.
Scaling such experience across scientific codebases requires accounting
for simulation cost, numerical variation, and multiple valid outcomes.
A verifier must specify which scientific quantities to compare, over
which workloads and tolerances, and how partial progress becomes reward.
These requirements extend the environment-scaling agenda developed for
code, terminals, browsers, and games
\citep{env-scaling-survey, swe-universe}.

\paragraph{Positioning ScienceIDE.}
Our comparison centers on task source, scientific verification, reuse,
and the learning interface. Benchmark instances provide a unit of evaluation;
ScienceIDE instead makes an expert-approved scientific module and its checks
a reusable unit for producing task instances. Environment-specific factories
adapt authoring procedures to that unit, and evaluation, SFT, and RL consume
the resulting interactions. The contribution is this connection between
scientific curation, repeated task production, and model learning, with
scientific content retained across changes in task family or trainer.

%% file: limitations.tex
\FloatBarrier
\section{Limitations}
\label{sec:limitations}

The evidence concerns reference-verifiable software tasks, predominantly
repair and implementation in computational physics and geoscience, not
open-ended discovery. Broader task-family coverage and the 1,000-environment
release remain development targets. Factory-generated variants can share code
paths and assumptions, so task count alone does not measure independent
scientific coverage. The hard set probes horizon and scale only up to a one-hour budget and two coupled edit sites.

Verification is a scientific modeling choice. Agreement with selected
observables and calibrated tolerances does not establish correctness
outside those checks, and legitimate alternatives may disagree with the
reference. Independent external audits and adversarial reward-hacking
evaluations remain outstanding. Factories amortize rather than eliminate
expert curation; expansion still depends on domain expertise and upstream
redistribution rights. Changes to repositories, dependencies, or hardware
require revalidation: versioning preserves provenance, not scientific validity.

Scores compare model--harness systems under fixed budgets; serving speed and
unequal repeat coverage limit fine-grained ranking. Retrieval controls cannot
exclude pretrained knowledge of public source code. Legacy images retain file-timestamp cues and difficulty selection predates complete retrieval isolation, so selection is not shown to be contamination-free.

SFT gains concern selected public benchmarks; RL gains concern hinted
held-out tasks within training environments. Neither establishes transfer to
unseen codebases. SFT separation is verified by task identifier, not related
variants; RL comparisons do not estimate between-seed variance.

%% file: conclusion.tex
\section{Conclusion}

ScienceIDE makes scientific experience reusable by turning expert judgments
into executable environments and checks.
Experiments expose a gap between plausible code and verified scientific
execution, widest where a task needs coordinated edits across a large
codebase or a numerical convention the code does not spell out, alongside selected public-benchmark SFT gains and hinted,
within-environment RL gains. The central design separates scientific judgment
from repeated task production: expert decisions become reusable contracts, and candidate tasks still require executable evidence.

The next step is to broaden scientific coverage and test transfer across
codebases and task families. Open-ended research also requires
checks that accommodate competing hypotheses and novel outcomes beyond a
fixed reference. The aim is a sustained exchange: science supplies grounded
experience for learning, and better agents extend the scientific work from
which future experience is built.

\clearpage

%% file: organizations.tex
\section{Organizations}
\label{app:organizations}

{\sffamily\mdseries\fontsize{10.5}{14}\selectfont\color{phaiink}
\setlength{\columnsep}{20pt}
\begin{multicols}{2}
\raggedright
\noindent\afl{1}{AItonomy Foundation}\par
\noindent\afl{2}{University of Oxford}\par
\noindent\afl{3}{University of California, Los Angeles}\par
\noindent\afl{4}{University of Texas at Austin}\par
\noindent\afl{5}{Princeton University}\par
\noindent\afl{6}{University of California, Berkeley}\par
\noindent\afl{7}{University of Michigan}\par
\noindent\afl{8}{California Institute of Technology}\par
\noindent\afl{9}{Boston University}\par
\noindent\afl{10}{Stanford University}\par
\noindent\afl{11}{University of California, San Diego}\par
\noindent\afl{12}{UNSW Sydney}\par
\noindent\afl{13}{University of California, Santa Cruz}\par
\noindent\afl{14}{Boston College}\par
\noindent\afl{15}{The Ohio State University}\par
\noindent\afl{16}{Asia Pacific Center for Theoretical Physics}\par
\noindent\afl{17}{Washington University in St. Louis}\par
\noindent\afl{18}{Cornell University}\par
\noindent\afl{19}{University of Illinois Urbana-Champaign}\par
\noindent\afl{20}{University at Buffalo}\par
\noindent\afl{21}{New Jersey Institute of Technology}\par
\noindent\afl{22}{New York University}\par
\noindent\afl{23}{University of California, Irvine}\par
\noindent\afl{24}{5Y Capital}\par
\noindent\afl{25}{Georgia Institute of Technology}\par
\noindent\afl{26}{Harvard University}\par
\noindent\afl{27}{Rutgers University}\par
\noindent\afl{28}{University of Maryland, College Park}\par
\noindent\afl{29}{Carnegie Mellon University}\par
\end{multicols}
}

%% file: appendix_compact_science.tex
\section{Scientific contracts and experience inventory}
\label{app:task-construction-protocols}

\subsection{Scientific check calibration}

A check fixes its inputs, graded outputs, and a pass policy.  Pointwise is
preferred when measured sensitivity fits a meaningful window and the bound
distinguishes a scientifically relevant deviation.  Outputs whose storage
order may change are aligned by an identity carried by the output; timings,
adaptive-step counts, rank layout, and random draws are not observables.
Invariants are used when pointwise comparison cannot contain valid variation
without losing discrimination.

Nominal and variant initial conditions provide calibration evidence: each run
is executed independently, must reproduce its reference and attain full
reward, and contributes a measured spread.  Where the build permits it, an
optional alternative legitimate build of the same pinned source records a
cross-build separation.  These observations are finite evidence, not a
universal floor or an automatic tolerance.  The curator and domain expert
choose the policy, bound, window, and variant by reading the source mechanism
and the scientific meaning of the observable.  Each check carries a warrant
that states what is compared, which scientifically relevant deviations the
bound is intended to distinguish, and why valid implementations can satisfy
it.  Fresh review can revise the check before acceptance.

Once fixed, the check suite is shared by repair, implementation, acceleration,
and reproduction families.  A factory may vary the agent's initial state,
deliverable, and task-specific objective, but it does not silently redefine
scientific equivalence.  Thus the expert work invested in a module's checks
is amortized across task families while acceptance remains executable.

\subsection{Cohort-bound experience inventory}
\label{app:experience-inventory}

The inventory keeps denominators attached to the population that produced
them.  The agent-evaluation cohort, scientific-contract cohort, and task-supply
cohort support different claims and must not be pooled: task supply and check
coverage do not change the ScienceIDE-Hard evaluation denominator.

\begin{table}[!htbp]
  \centering
  \small
  \caption{\textbf{Cohorts kept separate in the experience inventory.}}
  \label{tab:experience-cohorts}
  \begin{tabular}{p{0.23\linewidth}p{0.27\linewidth}p{0.38\linewidth}}
    \toprule
    Cohort & Scope & Interpretation \\
    \midrule
    Agent evaluation & 85 tasks across 18 environments &
      Fixed denominator for ScienceIDE-Hard model comparison. \\
    Scientific contracts & 1,076 executable checks &
      Broader observable-equivalence and invariant coverage. \\
    Task supply & 2,515 repair, 295 implementation, 2 acceleration &
      Manufactured task inventory, not an evaluation denominator. \\
    \bottomrule
  \end{tabular}
\end{table}

The table is an accounting boundary, not a claim that every environment has
the same task families or check density.  Detailed source records and links
remain in the Environment sources appendix.

\paragraph{Probing accounting.}
The policy-era escalation used one reference probe first and retried failures
up to three attempts: 2,722 attempts versus 4,845 under a uniform three-probe
policy, or 43.8\% fewer.  Across the pooled campaign, 3,195 graded
trajectories cost an estimated \$703 (about \$0.22 per trajectory), a
subscription-metered probe cost rather than an end-to-end construction cost.
These comparisons are descriptive, not preregistered.  The rate of
mixed-outcome tasks missed by single-success screening remains unaudited.

%% file: appendix_compact_evaluation.tex
\section{Evaluation protocol and integrity}
\label{app:agent-evaluation-protocol}

\paragraph{Fixed scientific cohort.}
ScienceIDE-Hard contains 85 tasks from 18 environments and five repository
families: 52 repair and 33 implementation tasks (MITgcm 33, PLUTO 31, LAPS
11, Athena++ 9, PHANTOM 1).  Every leaderboard model receives the same task
identifiers.  Each admitted task has a known-valid witness, empty-delivery
baseline, and measured defective starting point; strict success requires the
private scientific checks, not merely compilation or execution.

\paragraph{Harnesses, budget, and repeats.}
Agents use Codex, Claude Code, or Gemini CLI with identical task containers and
instructions, no localization hints, and a one-hour budget; results remain
model--harness measurements.  The analysis contains 3,982 validity-filtered
attempts. Fourteen models have at least three valid attempts on every task;
their point estimates average all observed valid repeats within each task
and then weight tasks equally. Fable uses its latest valid attempt on each
task and has no repeat interval. Intervals describe within-task repeat
variability. Budget curves use 1,200 seeded bootstrap replicates, drawing
three observations with replacement within each fixed task for the fourteen
repeated models; they are retrospective, not shorter-budget reruns.
Repeat-instability statistics use only the first three valid attempts per
task, separately from the point estimates.

For model $m$, the budget profile retained in the full analysis is
\begin{equation}
 \widehat p_m(b)=\frac{1}{N}\sum_{t=1}^{N}\frac{1}{K_{mt}}
 \sum_{i=1}^{K_{mt}}\mathbf{1}[S_{mti}=1,\,T_{mti}\le b],\quad N=85,
 \label{eq:budget-profile}
\end{equation}
where $K_{mt}$ is the selected repeat count (one for Fable and all observed
valid repeats for the other models) and $T$ is recorded duration capped at
one hour; this is not a shorter-budget rerun.

\subsection{Validity and execution-integrity audits}
\label{app:audits}

The reference-probe cohort contains 1,858 tasks across 25 environments and
five repository families; it files tiers but does not define the 85-task
comparison denominator.

\paragraph{Anchors, floors, and infrastructure errors.}
\label{app:gateaudit}
The reported core is validity-filtered, with oracle/witness, empty-delivery,
and defective-floor anchors checked.  The earlier bank audit was post hoc:
verifier crashes, missing references, OOMs, malformed rewards, and grader
errors are infrastructure errors excluded from the denominator, whereas a
valid \texttt{AgentTimeoutError} is budget exhaustion and counts as failure.
The historical census found 1,705 effective tasks, 136 without headroom, and
322 with unverified floors; this conditional audit is not certification of the
current full registry and does not replace the 85-task denominator.

\paragraph{Container-side retrieval.}
\label{app:netaudit}
The original campaign declared network isolation and an action-anchored fetch
audit, but later trajectory inspection found 455 upstream-directed commands
among 5,314 trials, with 21 confirmed successful fetches and 13 scored as
solved.  Nineteen staging scripts had rewritten the agent-side declaration to
\texttt{public}; the repository task files therefore did not describe the
containers that actually ran.  These are historical contamination findings,
not evidence that the earlier campaign was controlled.  The corrected path
uses a live-container allowlist and voids confirmed fetch trials.

\subsection{Provider and credential channels}
\label{app:provider-retrieval}
Container isolation does not cover provider-side search/fetch or an
aggregator credential that can call another browsing-capable model.  Gemini
received substantive provider-tool content on 71 of 85 tasks; one trial
retrieved an exact 22,180-character upstream file through \texttt{web\_fetch},
and affected trials were rerun with provider tools disabled.
DeepSeek V4 Pro used an aggregator key to call other models on 20 tasks,
solving 13; a provider-side model allowlist was required before rerunning.  In
that historical correction, the sample score moved from $.365$ to $.294$; this
is not a timeless current leaderboard value.  These fixes close the applicable
channel but do not claim pretrained models never saw public upstream code.

\subsection{Effort accounting and numerical details}
\label{app:efficiency}
\label{app:evaluation-tables}

Cost analyses use recorded harness estimates for nine models and archived
input/cached-input/output rates for six; cached input is subtracted first.
Two missing token totals came only from final structured usage records, and
missing native costs are not inferred.  Detailed leaderboard, budget,
efficiency, census, and answer-path tables remain frozen source artifacts.
Historical difficulty exploration establishes only that zero can mean a broken
task, budget exhaustion, retrieval, reward misalignment, or solver failure;
these are not standalone causal labels.

%% file: agent_diagnostics_appendix.tex
\section{Agent error analysis and scientific coverage}
\label{app:agent-diagnostics}
\label{app:trace-review}

This appendix complements the aggregate success, budget, and cost results
with failure behavior, scientific specialization, and trajectory-level
evidence on ScienceIDE-Hard.

\subsection{Failure outcomes and repeat instability}

\begin{figure}[!htbp]
  \centering
  \includegraphics[width=\linewidth]{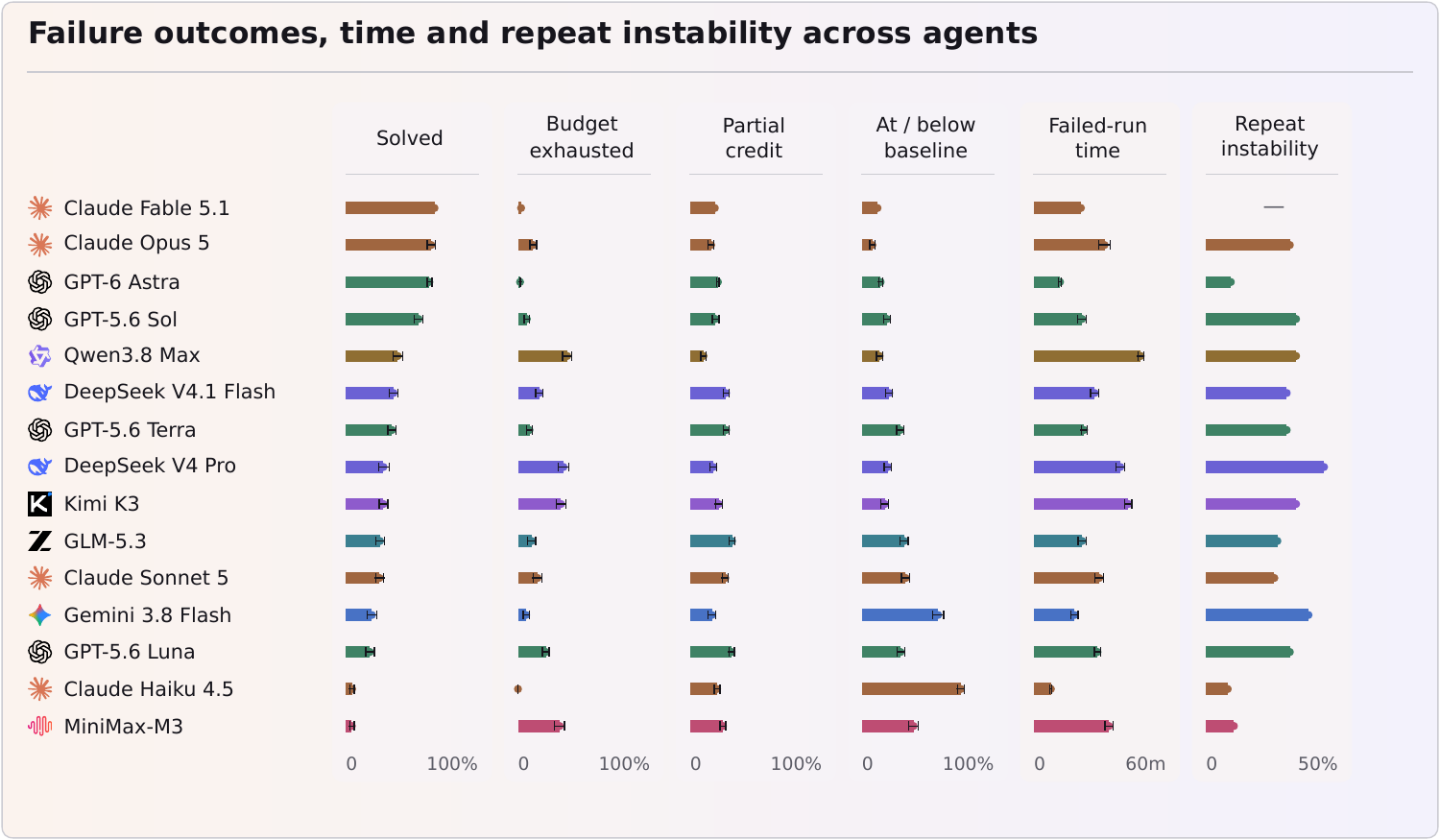}
  \caption{\textbf{Failure behavior and repeat instability are distinct diagnostics.}
  Task-balanced shares of success, budget exhaustion, non-timeout partial
  credit, and baseline-level returns, with mean unsuccessful-episode
  duration and first-three-repeat outcome instability. Error bars require
  complete repeat coverage. These are observable outcomes, not inferred
  causes of reasoning failure.}
  \label{fig:failure}
\end{figure}

\paragraph{Low success can reflect very different execution behavior.}
Qwen exhausts the budget on 37.3\% of its selected attempts and spends
48.5 minutes on an average unsuccessful episode. Haiku has no recorded
budget timeouts, averages 7.7 minutes on unsuccessful episodes, and
returns at or below baseline on 74.9\% of attempts.
MiniMax combines low success with 31.6\% budget exhaustion and
34.1-minute unsuccessful episodes (Figure~\ref{fig:failure}).
Failure can thus involve either brief unsuccessful work or prolonged
effort without a solution; termination status alone cannot establish why
an agent stopped.

Success is also unstable across repeated execution. Among agents with
complete coverage, 28.0\% of model--task pairs contain both success and
failure within their first three valid attempts. Gemini changes outcome
on 33 of 85 tasks, compared with eight for Astra and 29 for Kimi.
An aggregate success rate can therefore conceal unreliable task-level
behavior. The trajectory analysis below examines what differs between
successful and unsuccessful attempts.

\FloatBarrier

\subsection{Scientific specialization and complementarity}

\begin{figure}[!htbp]
  \centering
  \includegraphics[width=\linewidth]{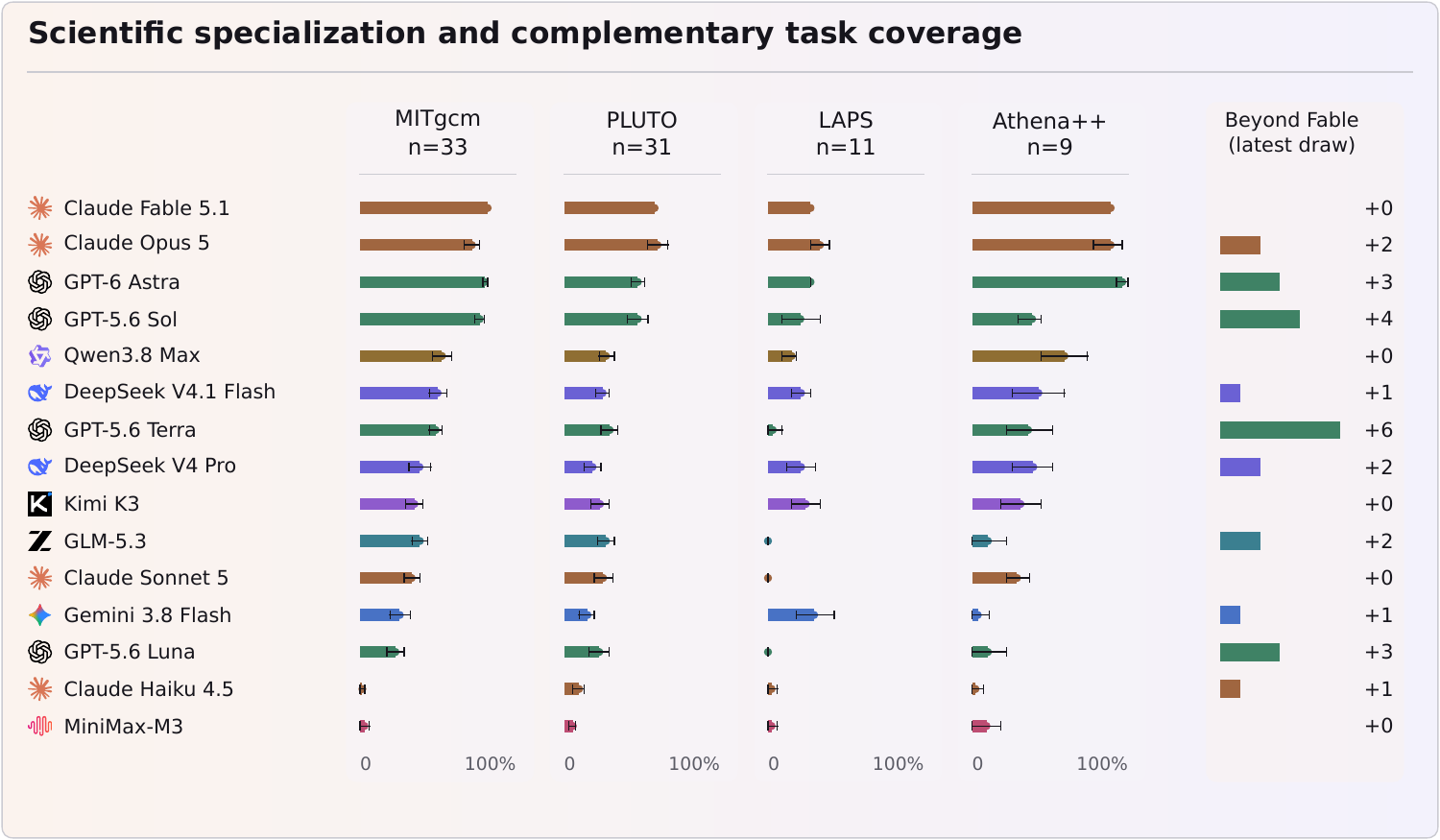}
  \caption{\textbf{Aggregate rank does not summarize scientific coverage.}
  Task-balanced success by repository family and additional successes
  beyond Fable in a common single-attempt sample. Error bars summarize
  repeated execution where coverage is complete. PHANTOM's single task
  is excluded from the family panels but retained in the 85-task coverage
  analysis. Additional solved sets may overlap.}
  \label{fig:family}
\end{figure}

\paragraph{Scientific capabilities are uneven and complementary.}
Astra and Sol have nearly equal PLUTO success rates (47.3\%), but differ
substantially on Athena++ (96.3\% versus 38.5\%).
Gemini's LAPS rate (29.5\%) exceeds Sol's (20.8\%), despite their opposite
aggregate ordering (Figure~\ref{fig:family}).
These reversals reveal uneven performance across scientific software;
the families also differ in language and task composition, so the
comparison does not isolate a domain effect.

In the common single-attempt coverage sample, Fable solves 57 of 85
tasks, while the union across all fifteen agents solves 72.
Terra contributes six successes outside Fable's solved set despite its
lower aggregate score. Eleven tasks are solved by exactly one agent, and
13 by none. The union indicates observed complementarity, not the
performance of a tested ensemble: realizing it would require a selection
or verification policy and additional computation.
\FloatBarrier

\subsection{Trajectory-level failure mechanisms}
\label{sec:failure-mechanisms}
\label{app:trace-contrasts}
\label{app:trace-case-evidence}

\input{trajectory_failures}
\FloatBarrier

\subsection{Trajectory review protocol}
\label{app:trace-review-protocol}

We reviewed Fable~5.1 and Astra on the common 85-task panel, covering 358
attempts and 125 unsuccessful attempts, including successful repeats as
within-task contrasts. A model-assisted reviewer read the original actions,
tool outputs, and grading records. The annotations are evidence-supported
mechanism labels, not experimentally isolated causes: the review was not
blinded, independently replicated, or domain-expert certified, and 17 attempts
remained unresolved. The task source files were not independently proven
byte-identical to every historical revision, so code-level attributions retain
that provenance limitation. No replayed command was executed.

The review is task-balanced rather than attempt-weighted. Repeated local tests
show that an agent's chosen edit is reproducible, but they do not establish
that it addresses the assigned defect; private-reference grading remains the
acceptance test. Detailed task-level records and evidence references remain in the
accompanying source artifacts.

%% file: trajectory_failures.tex
We examine why agents fail by reviewing Fable 5.1 and Astra on all
85 ScienceIDE-Hard tasks, including successful repeats as within-task
contrasts. The review covers 358 attempts, of which 125 are unsuccessful.
A model-assisted reviewer inspects original actions, tool outputs, and
grading records for each task. The resulting annotations describe
evidence-supported mechanisms, not experimentally isolated causes;
the protocol is in Appendix~\ref{app:trace-review-protocol}, and
representative cases are presented below.

\begin{figure}[!t]
  \centering
  \includegraphics[width=\linewidth]{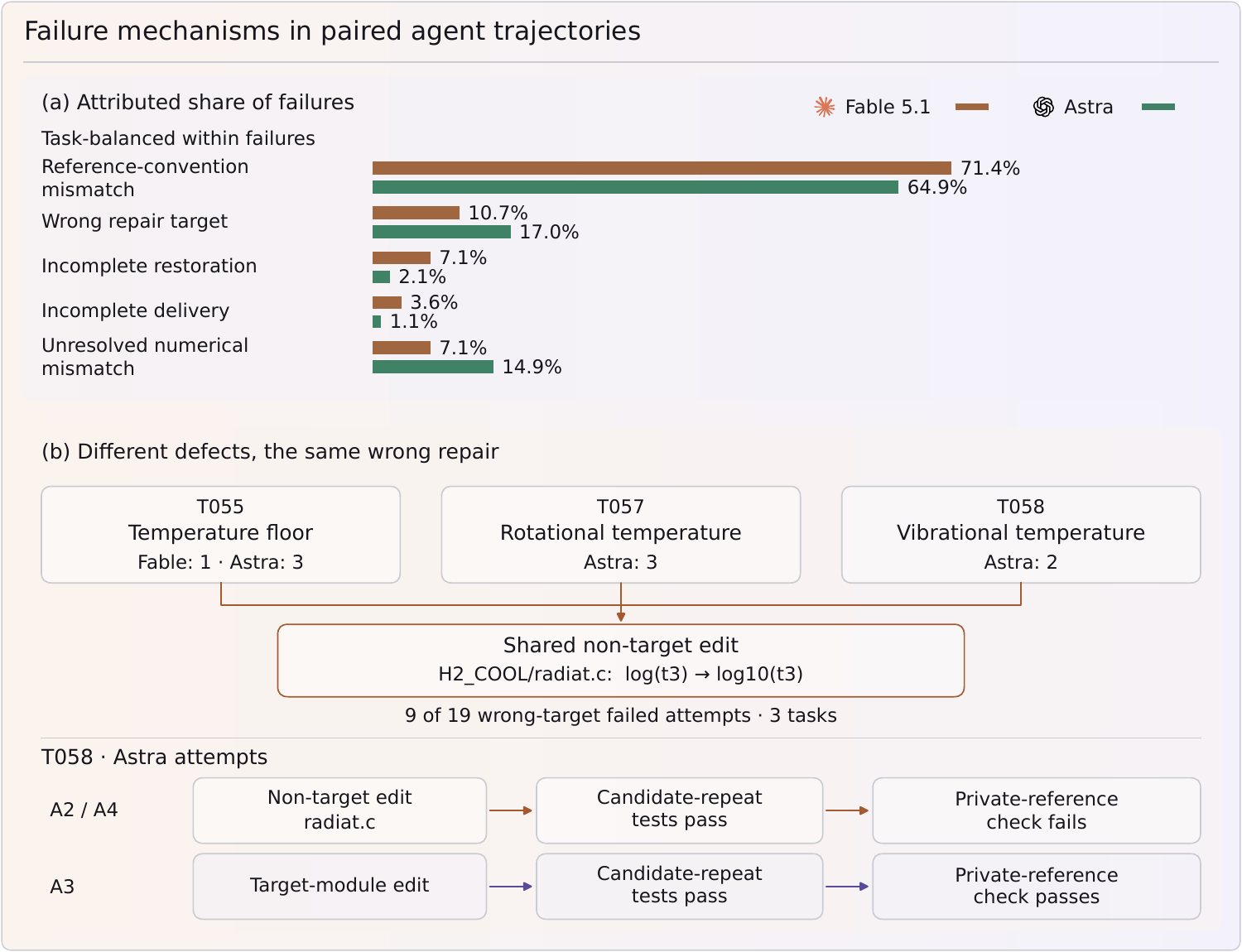}
  \caption{\textbf{Different defects can lead to the same non-target repair.}
  (a) Primary mechanism annotations, task-balanced within unsuccessful attempts.
  (b) Nine failed attempts across three tasks converge on one non-target edit;
  a same-task Astra success repairs the intended target.
  Candidate-repeat checks pass on both paths, but private-reference outcomes
  differ. A2--A4 identify individual attempts.}
  \label{fig:trace-causes}
\end{figure}

\paragraph{Scientific reconstruction requires codebase-specific conventions.}
Reference-convention mismatches account for 71.4\% of Fable's and 64.9\%
of Astra's task-balanced failures (Figure~\ref{fig:trace-causes}a).
In a LAPS timestep task (T013), both models produce runnable code but omit
the rule for retaining the previous timestep. The lower-resolution case
matches the reference, while the higher-resolution case diverges at a
periodic update. In an MITgcm tracer task (T035), all four attempts
reconstruct nearly the same coefficient table with one truncated coefficient;
the associated tracer diverges while another matches.
The challenge is therefore not only recovering a scientific formula,
but reproducing the numerical precision and state conventions required
by the surrounding code.

\paragraph{Different defects can induce the same wrong-target repair.}
Wrong-target edits account for 10.7\% and 17.0\% of the two models'
task-balanced failures. Nine failed attempts across three PLUTO cooling
tasks make the same natural-log to base-10-log change in a non-target
routine, although the assigned defects concern three different quantities:
a temperature floor (T055), rotational temperature (T057), and vibrational
temperature (T058). The intended defects remain unrepaired.
On T058, a successful Astra repeat instead restores the target constant
in the equation-of-state module (Figure~\ref{fig:trace-causes}b).
This contrast identifies a recurring localization error rather than an
isolated implementation typo; it does not establish why the non-target
expression attracts the models.

\paragraph{A passing local test may not test the assigned task.}
Both the unsuccessful and successful T058 repairs pass comparisons between
repeated runs of their own candidate code. Only the target repair passes
the private-reference evaluation. Similarly, failed Astra attempts on a
PLUTO substep-handoff task (T059) test a modified error-clamp routine while
leaving the defective handoff unchanged; the successful repeat repairs
the handoff. The local tests establish repeatability or behavior of the
chosen edit, not its relevance to the task.
All three Astra attempts on T057 repeat the same non-target logarithm edit,
and all three on T013 omit the same timestep-retention rule. Repeated
execution can thus reproduce a specific mistake, not just independent
failures with different explanations.

\paragraph{Correct code does not guarantee a complete scientific delivery.}
On an Athena++ chemistry task (T002), Fable and two Astra attempts apply
the correct target edit, but omit four required control-output groups.
Four other Astra attempts deliver the complete run set and pass.
The failure is therefore incomplete scientific execution rather than an
incorrect repair. Together with the localization contrasts, this case
separates three requirements: selecting the relevant change, reconstructing
the required behavior, and completing the deliverables.
All eight mixed-outcome Astra tasks were included in the review; the cases
above illustrate the observed mechanisms. Task-level annotations are
provided in the accompanying source artifact
\texttt{data/trace\_failure\_review.json}.

Seventeen unsuccessful attempts remain unresolved after targeted re-review,
often because plausible numerical reconstructions still disagree with the
reference. The annotations are neither expert-certified causal labels nor
evidence that the verifier is wrong. For learning from these trajectories,
the useful distinction is between the edit, what its local tests establish,
the delivered artifacts, and the task-level outcome. Localization and
delivery failures suggest different feedback; unresolved cases should
retain their uncertainty.

%% file: appendix_compact_learning.tex
\section{Learning settings and supplementary results}

\subsection{Supervised fine-tuning}
\label{app:sft-training-details}

Qwen3.5-4B, Qwen3.5-9B, and Qwen2.5-72B-Instruct were trained with
ms-swift on GPT-5.6-sol demonstrations selected by numerical-equivalence
verification. The task-identifier-disjoint corpus contains 4,567 training
segments from 564 tasks and 544 validation segments from 81 tasks.
The loss supervises new assistant actions and tool calls; instructions,
observations, reused context, and flagged repetitive or failed actions
are masked. Related variants were not independently audited for overlap.

\paragraph{Example construction.}
Each example consists of the initial instruction, bounded interaction
history, and a new target segment ending at a complete
assistant--observation block. The loss supervises selected assistant
tokens, including tool calls, in the new segment. Instructions,
observations, and reused history are masked. Tool-action targets identified as
failed, repetitive, or part of an extended read-only sequence are
also masked, with their messages retained as context. Successive
examples can thus share context without repeating supervised targets.

LoRA targets all linear layers with rank 32, scaling 64, and dropout
0.05. Runs use BF16, global batch size 64, a 36,864-token sequence limit,
learning rate $2\times10^{-5}$, 3\% warmup, cosine decay, and three epochs
(216 steps). Final checkpoints are evaluated without selecting them
on public-benchmark performance. Table~\ref{tab:sft-training-diagnostics}
reports teacher-forced validation loss and token accuracy on unmasked
targets.

\begin{table}[!htbp]
\centering
\caption{\textbf{SFT validation diagnostics.} Loss is measured after
epochs one and three; final token accuracy uses unmasked validation
targets.}
\label{tab:sft-training-diagnostics}
\small
\begin{tabular}{lrrr}
\toprule
Model & Loss, epoch 1 & Loss, epoch 3 & Final acc.\ (\%) \\
\midrule
Qwen3.5-4B & 0.383 & 0.340 & 91.12 \\
Qwen3.5-9B & 0.369 & 0.325 & 91.33 \\
Qwen2.5-72B-Instruct & 0.401 & 0.313 & 92.01 \\
\bottomrule
\end{tabular}
\end{table}

\paragraph{Evaluation and confirmation.}
The three-model study screens 33 benchmark configurations with 20--128
items each. Multiple-choice configurations use continuation likelihoods or
option extraction from generated responses. Other generated answers are scored
by numerical or symbolic comparison, exact match, or execution
against frozen tests. Generation is greedy, with thinking disabled
and a 4,096-token output limit. Within each model pair, both checkpoints use the
initial model's tokenizer and the same context limit:
65,536 tokens for 4B/9B and 32,768 for 72B. Selected findings are evaluated on disjoint confirmation items.
A separate 4B/9B extension covers 21 further configurations.
Reported intervals are paired-bootstrap intervals without correction
for multiple comparisons.

\paragraph{Screening gains by benchmark.}
Beyond the largest code gains reported in \S\ref{sec:sftresults}, the
remaining code comparisons show smaller gains. HumanEvalFix Rust improves
by 3.66 points for 9B. For 72B, APPS Introductory improves by 6.25
points and LiveCodeBench Execution by 5.47 points, while HumanEvalFix
C++ and HumanEval gain 4.69 and 3.125 points, respectively.
On QuixBugs Python, both 4B and 72B improve by 5.0 points,
from 0.600 to 0.650 and from 0.900 to 0.950, respectively.
The same absolute gain therefore occurs at different initial
performance levels.

The improvements also extend to reasoning and knowledge benchmarks.
For 9B, BBH Word Sorting increases from 0.240 to 0.576
(+33.6 points), alongside a 3.125-point gain on GSM8K.
For 4B, BBH multistep arithmetic increases from 0.920 to 0.968
(+4.8 points), and MMLU-Pro Computer Science gains 4.69 points.
ARC-Easy improves by 3.125 points for 72B.

\input{sft_confirmation_table}

On the 250-item BBH full splits, 9B Word Sorting improves from 25.6\%
to 60.4\% and 4B multistep arithmetic from 92.4\% to 97.6\%.
Their respective 125-item confirmation comparisons are
27.2\% to 63.2\% and 92.8\% to 98.4\%.
Full splits combine screening and confirmation and are not additional
independent replications. The 72B HumanEvalFix C++ screening gain does
not persist on confirmation. Machine-readable confirmation and
screening records are retained in the accompanying data files;
Figure~\ref{fig:sft-benchmark-results} does not summarize all
benchmarks as uniformly improved.

\subsection{Reinforcement learning}
\label{app:rl-implementation}

RL starts from base Qwen3.5-4B without SFT initialization.
LAPS uses 99 tasks split 85/14. The MITgcm-biogeo run records 87 tasks
with a source split of 64/23, while its logged validation reads 21 tasks.
A rebuilt split is 66/21 and is not substituted for the reported run.
Factory splits group defect-family and source-tree variants.
Both training and validation use L1 hints identifying file, line, and
edit class without the repair, and the held-out comparison is within each
environment.

\paragraph{Execution stack.}
Every iteration rolls out episodes, grades them, and takes one optimizer
step (Figure~\ref{fig:rl_overview} in \S\ref{sec:rlresults}). Generation runs in vLLM, optimization
in PSRL, a customization of the veRL trainer that drives a black-box
harness, and episodes execute in harbor through the same Terminus-2
containers and verifiers as the evaluation campaign. This design lets
training consume the verifier's native reward rather than a learned proxy,
since the trainer does not need to model the execution inside each
container. The resulting episodes are nevertheless expensive and variable
in length. Rollout and training therefore occupy disjoint GPUs and run
concurrently, allowing the training workers to make progress while new
episodes are being evaluated.

\paragraph{Optimization and execution.}
The algorithmic choices are motivated in \S\ref{sec:rlresults}. This
section records the settings behind them and the run-level evidence for
the truncation mask.
Episode execution uses \texttt{TruncatingTerminus2} with 16 concurrent
episodes, a 2,700-second task timeout, a 900-second verifier timeout, and
4,096-byte observation truncation. Each run uses 24 H20 GPUs
with 95GB per device: eight for generation (tensor parallelism 2,
four instances) and sixteen for training (FSDP2, sequence parallelism 4).
On a representative step the optimizer update took 2,069 of 3,210 seconds
and rollout 751 seconds, which is the utilization motivating disjoint
rollout and training pools. Trajectory capture is token-in-token-out:
each turn's message prefix is hashed so that turn $k+1$ extends turn $k$,
and reasoning is retained across turns so every turn is a strict prefix
extension. A stock chat template that strips the previous turn's reasoning
would miss the prefix hash and reopen a trajectory per turn, making
storage and backward cost grow with the square of the turn count.
Table~\ref{tab:compact-learning-setup} records the optimization settings.

\begin{table}[!htbp]
\centering
\small
\setlength{\tabcolsep}{4pt}
\caption{Reinforcement-learning run settings, carried unchanged by both
reported runs. The left block lists the algorithmic choices motivated in
\S\ref{sec:rlresults}, the right block the remaining numerical settings.
The advantage baseline follows Dr.~GRPO \citep{drgrpo}, the asymmetric clip
follows DAPO Clip-Higher \citep{dapo}, and rollout correction is token-level
truncated importance sampling.}
\label{tab:compact-learning-setup}
\begin{tabular}{@{}ll@{\hspace{1.25em}}ll@{}}
\toprule
Algorithmic choice & Value & Run setting & Value \\
\midrule
Advantage baseline & Group mean, no std.  & Batch / rollouts     & 16 / 8 \\
Clipping           & $0.2$ / $0.3$        & Learning rate        & $10^{-6}$ \\
Loss aggregation   & Token mean           & Warmup / weight decay & 3 steps / 0.1 \\
Truncated episodes & Masked from loss     & Turn limit           & 50 \\
Length penalty     & Off                  & Prompt / response    & 2,048 / 65,536 \\
Rollout correction & Cap $2.0$            & Temperature          & 1.0 \\
Dynamic sampling   & Off                  & Staleness            & 1 update \\
KL / entropy bonus & Off / off            & Measured KL          & $6\times10^{-4}$ \\
                   &                      & Generation / trainer & vLLM / veRL \\
\bottomrule
\end{tabular}
\end{table}

The advantage of Eq.~\eqref{eq:rl-adv} omits the standard-deviation
normalization of the original GRPO formulation. Verifier rewards in our setting
are close to bimodal, with most groups scoring nearly all zero or nearly all
one. Dividing by a small standard deviation would therefore turn an isolated
success into a disproportionately large advantage.
With this advantage, each token of trajectory $i$ contributes
\begin{equation}
  \ell_{i,t}(\theta) = \min\!\big(r_{i,t}\hat{A}_i,\;
    \mathrm{clip}(r_{i,t},\,1-\varepsilon_{\mathrm{low}},\,
    1+\varepsilon_{\mathrm{high}})\,\hat{A}_i\big),
  \label{eq:rl-clip}
\end{equation}
where $r_{i,t}$ is the per-token probability ratio against the sampling
policy. The bound is asymmetric, with $\varepsilon_{\mathrm{low}}=0.2$ and
$\varepsilon_{\mathrm{high}}=0.3$, so that rarely sampled repair actions
retain room to gain probability mass \citep{dapo}, and the loss is
aggregated by token mean.
Rollout correction uses token-level truncated importance sampling to account
for policy staleness, which arises because a step trains on trajectories from
the previous weights and because rollout runs in BFloat16 while training uses
higher precision. The weight is $w_{i,t}=\min(\exp \rho_{i,t},\,C)$ with
$C=2.0$, where $\rho_{i,t}$ is the trainer-minus-rollout log-probability
difference for that token, computed in log space. SFT and RL are reported
as independent studies.

\paragraph{Termination handling and diagnostics.}
Budget-truncated trajectories retain their reward contribution to the
group baseline but have no token-loss contribution:
\begin{equation}
  m_{i,t} \;=
  \begin{cases}
    0 & \text{if } \tau_i \in \mathcal{T}_{\text{budget}},\\
    \mathbb{1}[\text{token } t \text{ is model-generated}] & \text{otherwise},
  \end{cases}
  \label{eq:rl-mask}
\end{equation}
The implementation applies the mask after termination classification, so $\mathcal{T}_{\mathrm{budget}}$
contains turn- and response-budget cutoffs and excludes infrastructure
timeouts and completed grading errors. Prompt tokens and harness padding are
excluded from the mask. A batch in which every trajectory is budget-truncated
would give the token-mean denominator no support, so it is clamped to at
least one and such a batch yields a zero gradient.

The run-level comparison behind that choice is
Table~\ref{tab:rl-truncation}. Across the collapsed run, effective
sample size remains in $[0.950,0.978]$ and rollout/trainer
log-probability correlation recovers from $0.938$ to $0.955$ while reward
falls. These diagnostics argue against a large deterioration in
importance-sampling quality during the collapse.
Over the reported runs, effective sample size stays in
$[0.950,0.978]$ for LAPS and $[0.953,0.976]$ for MITgcm-biogeo, with
log-probability correlations in $[0.948,0.984]$ and $[0.965,0.977]$.
These bound weight transfer and policy mismatch. The audit of the
reward-side penalty we switch off found it firing on 28\% of samples over one
run, of which 119 of 142 were already-failing cases, and hitting finished
episodes scoring 1.0 about 64\% of the time. The fraction of groups with zero
reward variance, which contribute no gradient under Eq.~\eqref{eq:rl-adv},
grows from 10.0\% to 47.5\% over the LAPS run and from 27.5\% to 41.3\% over
MITgcm-biogeo, comparing the first and last five recorded steps.

\begin{table}[!htbp]
\centering
\caption{Run-level comparison with and without the truncation mask, from the
first to the last recorded step of each run, with the unmasked run shown
through its peak to its collapse. Tokens per turn and turns per episode
decompose response length.}
\label{tab:rl-truncation}
\small
\begin{tabular}{lcc}
\toprule
& Unmasked & Masked \\
\midrule
Mean reward       & $0.339 \to 0.573 \to 0.078$ & $0.487 \to 0.914$ \\
Tokens per turn   & $1125 \to 327$              & $1095 \to 1165$ (flat) \\
Truncation rate   & $29\% \to 39\%$             & $28.9\% \to 2.3\%$ \\
Turns per episode & $34.9 \to 49.5$             & $32.2 \to 24.1$ \\
Entropy           & Rising                      & $0.586 \to 0.348$ \\
\bottomrule
\end{tabular}
\end{table}

Comparing the first and last five recorded steps of each reported run,
training reward rises from $0.427$ to $0.828$ on LAPS and from $0.381$ to
$0.597$ on MITgcm-biogeo, budget truncation falls from 39.5\% to 6.6\% and
from 34.1\% to 23.8\%, and entropy falls from $0.602$ to $0.381$ and from
$0.549$ to $0.313$. Turns per episode fall from $34.7$ to $24.5$ on LAPS while
tokens per turn hold near $1100$, so the shortening reflects earlier
completion instead of truncated turns.
\FloatBarrier

%% file: sft_confirmation_table.tex
\begin{table}[!htbp]
\centering
\small
\setlength{\tabcolsep}{3pt}
\caption{Confirmation comparisons shown in Figure~\ref{fig:sft-benchmark-results}(c).
Scores are percentages; changes and 95\% paired-bootstrap intervals
are percentage points. Rows retain their own accuracy, exact-match,
or execution metric. Intervals are unadjusted for multiple comparisons.}
\label{tab:sft-confirmation}
\begin{tabular}{@{}p{0.43\linewidth}rrrr@{}}
\toprule
Dataset / model ($n$) & Initial & SFT & Change & 95\% interval \\
\midrule
CodeXGLUE defect / 4B (2,604) & 45.93 & 52.92 & +6.99 & [4.03, 9.87] \\
CodeXGLUE refinement / 9B (5,707) & 1.17 & 2.37 & +1.19 & [0.89, 1.51] \\
CodeXGLUE refinement / 72B (5,707) & 1.63 & 2.40 & +0.77 & [0.53, 1.03] \\
BBH Word Sorting / 9B (125) & 27.20 & 63.20 & +36.00 & [26.40, 46.40] \\
BBH arithmetic / 4B (125) & 92.80 & 98.40 & +5.60 & [0.80, 11.20] \\
HumanEvalFix Python / 9B (36) & 86.11 & 77.78 & -8.33 & [-19.44, 2.78] \\
\bottomrule
\end{tabular}
\end{table}

%% file: environment_sources_compact.tex
\section{Environment sources}
\label{app:envsources}

The frozen registry cohort contains 27 upstream scientific codebases and 64
derived environments.  The list retains every source, its primary citation,
the pinned upstream repository link, and the license recorded in the codebase
report; the release ledger supplies the complete environment-to-source map.

{\small
\setlength{\parindent}{0pt}
\setlength{\parskip}{1pt}
\textbf{Athena++ astrophysical radiation hydrodynamics and MHD} \citep{Stone2020Athena}. \href{https://github.com/PrincetonUniversity/athena}{\texttt{athena}}, BSD-3-Clause.\par
\textbf{Basilisk spacecraft and balanced reaction-wheel dynamics} \citep{kenneally2020basilisk}. \href{https://github.com/AVSLab/basilisk}{\texttt{basilisk}}, ISC.\par
\textbf{dscribe} \citep{Himanen2020DScribe}. \href{https://github.com/SINGROUP/dscribe.git}{\texttt{dscribe}}, Apache-2.0.\par
\textbf{EDKit.jl} \citep{edkit2026}. \href{https://github.com/jayren3996/EDKit.jl}{\texttt{EDKit.jl}}, MIT.\par
\textbf{eftcamb} \citep{Hu2014EFTCAMB}. \href{https://github.com/EFTCAMB/EFTCAMB.git}{\texttt{eftcamb}}, GPL-3.0 (EFTCAMB part); CAMB licence for unmodified CAMB.\par
\textbf{EPOCH particle-in-cell code} \citep{Arber2015EPOCH}. \href{https://github.com/epochpic/epoch}{\texttt{epoch}}, GPL-3.0 (task metadata); GPL-3.0-or-later in inspected Fortran file headers.\par
\textbf{G4CMP} \citep{kelsey2023g4cmp}. \href{https://github.com/kelseymh/G4CMP.git}{\texttt{G4CMP.git}}, GPL-3.0-or-later (bundles Geant4 licence and Qhull).\par
\textbf{Gkeyll Computational Plasma Physics Package} \citep{gkeyll2026}. \href{https://github.com/gkeyllorg/gkeyll}{\texttt{gkeyll}}, MIT.\par
\textbf{LAPS: 3D and 2D compressible Hall-MHD} \citep{Shi2024LAPS}. \href{https://github.com/chenshihelio/LAPS}{\texttt{LAPS}}, GPL-2.0.\par
\textbf{Meep 1.34.0 — finite-difference time-domain electromagnetics} \citep{oskooi2010meep}. \href{https://github.com/NanoComp/meep}{\texttt{meep}}, GPL-2.0-or-later.\par
\textbf{Mink differential inverse kinematics} \citep{zakka2026mink}. \href{https://github.com/kevinzakka/mink}{\texttt{mink}}, Apache-2.0.\par
\textbf{MITgcm (MIT General Circulation Model)} \citep{Marshall1997FiniteVolume}. \href{https://github.com/MITgcm/MITgcm}{\texttt{MITgcm}}, MIT.\par
\textbf{NEST: Noble Element Simulation Technique} \citep{szydagis2011nest}. \href{https://github.com/NESTCollaboration/nest}{\texttt{nest}}, Apache-2.0.\par
\textbf{EPREM} \citep{Schwadron2010EMMREM}. \href{https://gitlab.com/open-eprem/eprem.git}{\texttt{eprem.git}}, GPL-3.0-only.\par
\textbf{Phantom} \citep{price2018phantom}. \href{https://github.com/danieljprice/phantom}{\texttt{phantom}}, GPL-3.0.\par
\textbf{PLUTO — Godunov-type computational astrophysical fluid dynamics} \citep{Mignone2007PLUTO}. \href{https://plutocode.ph.unito.it/}{\texttt{pluto}}, GPL-2.0.\par
\textbf{PyAMG} \citep{pyamg2023}. \href{https://github.com/pyamg/pyamg}{\texttt{pyamg}}, MIT.\par
\textbf{pymatgen} \citep{Ong2013Pymatgen}. \href{https://github.com/materialsproject/pymatgen}{\texttt{pymatgen}}, MIT.\par
\textbf{pymatgen-core} \citep{Ong2013Pymatgen}. \href{https://github.com/materialsproject/pymatgen-core}{\texttt{pymatgen-core}}, MIT.\par
\textbf{pyXSIM} \citep{zuhone2014xray}. \href{https://github.com/jzuhone/pyxsim}{\texttt{pyxsim}}, BSD-3-Clause.\par
\textbf{QuTiP — Quantum Toolbox in Python} \citep{lambert2026qutip5}. \href{https://github.com/qutip/qutip}{\texttt{qutip}}, BSD-3-Clause.\par
\textbf{S4: Stanford Stratified Structure Solver} \citep{liu2012s4}. \href{https://github.com/victorliu/S4}{\texttt{S4}}, GPL-2.0-or-later.\par
\textbf{Scirpy — single-cell immune receptor repertoire analysis} \citep{Sturm2020Scirpy}. \href{https://github.com/scverse/scirpy}{\texttt{scirpy}}, BSD-3-Clause.\par
\textbf{NASA SimuPy Flight: NESC 6-DoF dynamics} \citep{MargolisLyons2022SimuPyFlight}. \href{https://github.com/nasa/simupy-flight}{\texttt{simupy-flight}}, NASA-1.3.\par
\textbf{Stim — a fast stabilizer circuit simulator} \citep{gidney2021stim}. \href{https://github.com/quantumlib/Stim}{\texttt{Stim}}, Apache-2.0.\par
\textbf{strax: streaming analysis for xenon TPCs} \citep{Aalbers2026Strax223}. \href{https://github.com/AxFoundation/strax}{\texttt{strax}}, BSD-3-Clause.\par
\textbf{TSID TALOS fixed-contact whole-body inverse dynamics} \citep{delprete2016torquecontrol}. \href{https://github.com/stack-of-tasks/tsid}{\texttt{tsid}}, BSD-2-Clause.\par
}